\documentclass[10pt,twocolumn,letterpaper]{article}

\usepackage[pagenumbers]{iccv} 

\definecolor{iccvblue}{rgb}{0.21,0.49,0.74}
\usepackage[breaklinks,colorlinks,citecolor=iccvblue]{hyperref}

\usepackage{pifont}
\newcommand{\cmark}{\ding{52}}%
\newcommand{\xmark}{\ding{56}}%
\newcommand{\hcmark}{\ding{52}\rotatebox[origin=c]{-9.2}{\kern-0.7em\ding{55}}}
\usepackage{subcaption}
\usepackage[most]{tcolorbox}
\usepackage{colortbl}
\usepackage{graphicx}
\usepackage{capt-of} 

\usepackage{listings}

\definecolor{correctcolor}{RGB}{182,232,128}

\def\paperID{7472} 
\def\confName{ICCV}
\def\confYear{2025}

\title{MC-Bench: A Benchmark for Multi-Context\\ Visual Grounding in the Era of MLLMs}

\author{Yunqiu Xu\\
ReLER Lab, CCAI\\
Zhejiang University\\
{\tt\small imyunqiuxu@gmail.com}
\and
Linchao Zhu\footnotemark\\
ReLER Lab, CCAI\\
Zhejiang University\\
{\tt\small zhulinchao@zju.edu.cn}
\and
Yi Yang\\
ReLER Lab, CCAI\\
Zhejiang University\\
{\tt\small yangyics@zju.edu.cn}
}

\begin{document}

\twocolumn[{
\maketitle
\centering
\vspace{-2em}
\begin{minipage}{0.325\textwidth}
\centering
\includegraphics[width=\linewidth]{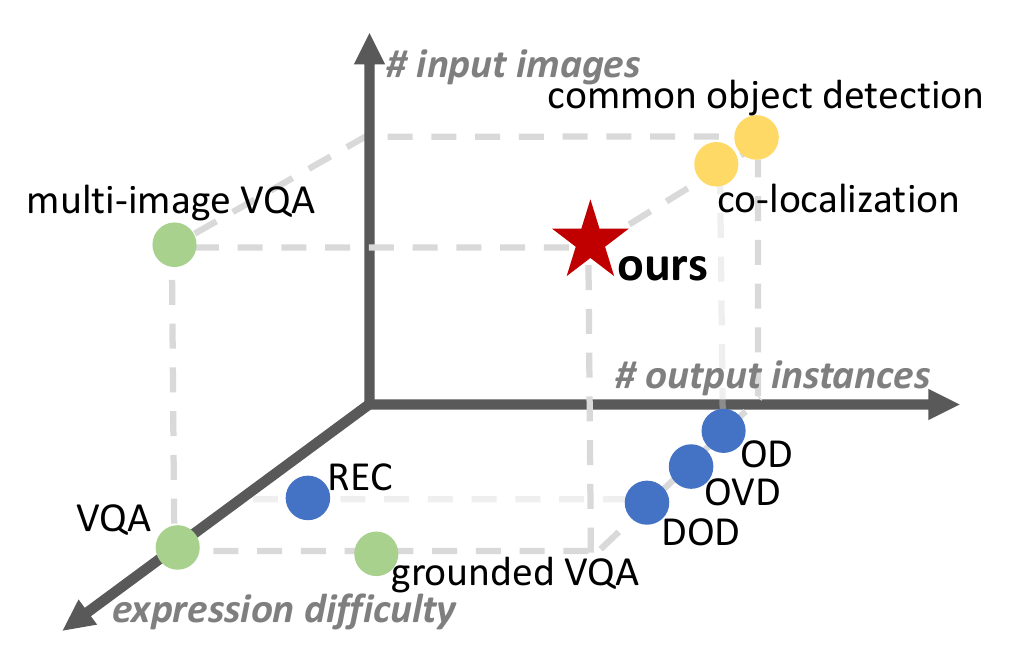}
\vspace{-1.85em}
\captionsetup{type=figure}
\subcaption{correlation of our proposed task with others}
\label{fig: task compare}
\end{minipage}
\begin{minipage}{0.325\textwidth}
\centering
\includegraphics[width=\linewidth]{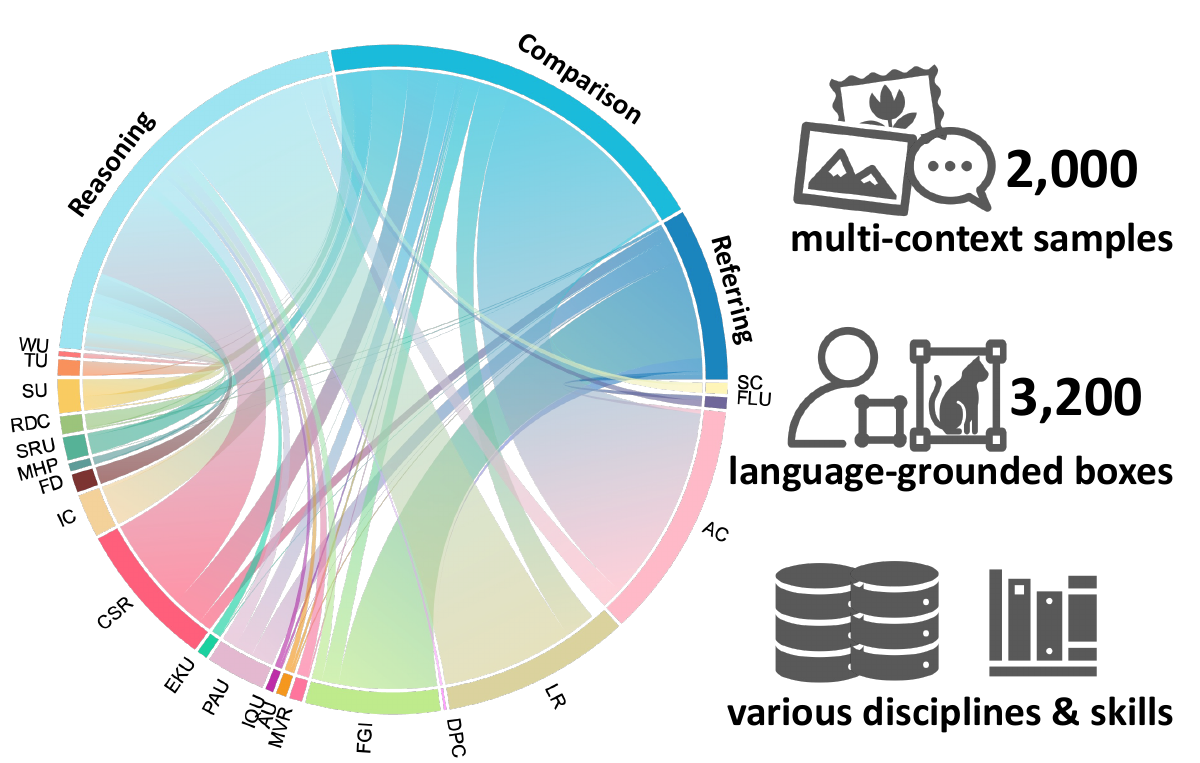}
\vspace{-1.55em}
\captionsetup{type=figure}
\ContinuedFloat
\subcaption{diverse images and texts covering various tasks}
\end{minipage}
\begin{minipage}{0.325\textwidth}
\centering
\includegraphics[width=\linewidth]{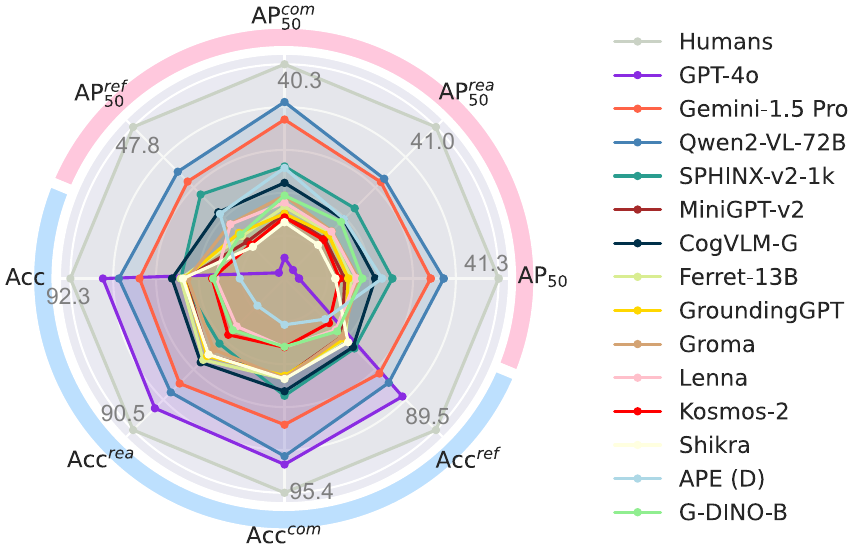}
\vspace{-1.7em}
\captionsetup{type=figure}
\ContinuedFloat
\subcaption{benchmark results of representative baselines}
\end{minipage}
\vspace{-0.75em}
\setcounter{figure}{0}
\captionof{figure}{Multi-context visual grounding is a new task that aims at localizing target instances based on open-ended text prompts in multi-image scenarios. A new dataset MC-Bench is constructed to benchmark the MLLMs and foundation models with potential multi-context visual grounding capabilities. The benchmark results of over 20 state-of-the-art models reveal a significant performance gap between existing approaches and humans, while also suggesting potential future directions.}
\label{fig:teaser}
\vspace{0.5em}
}]

\renewcommand{\thefootnote}{\fnsymbol{footnote}}
\footnotetext{*Corresponding author.}
\renewcommand{\thefootnote}{\arabic{footnote}}

\begin{abstract}
While multimodal large language models (MLLMs) have demonstrated extraordinary vision-language understanding capabilities, their abilities to solve instance-level visual-language problems beyond a single image warrant further exploration.
To assess these unproven abilities of MLLMs, this paper proposes a new visual grounding task called multi-context visual grounding, which aims to localize instances of interest across multiple images based on open-ended text prompts. 
In order to facilitate this research, we construct a new dataset MC-Bench that features 2K high-quality and manually annotated samples.
Each sample consists of an instance-level labeled image pair and a corresponding text prompt that indicates the target instances in the images. 
These text prompts are highly open-ended and follow three distinct styles, covering 20 practical skills.
We benchmark over 20 state-of-the-art MLLMs and foundation models with potential multi-context visual grounding capabilities, along with our developed simple yet effective agentic baseline and a finetuned baseline by multi-context instruction tuning.
Our evaluation reveals a non-trivial performance gap between existing MLLMs and humans, along with some insightful observations that suggest potential future directions.
We hope that MC-Bench and our empirical findings encourage the research community to further advance the untapped potentials of MLLMs in instance-level tasks, particularly in multi-image contexts.
Project page: \url{https://xuyunqiu.github.io/MC-Bench}.
\end{abstract}
\vspace{-1em}

\section{Introduction}\label{sec:intro}

\begin{figure*}[t]
\centering
\includegraphics[width=\linewidth]{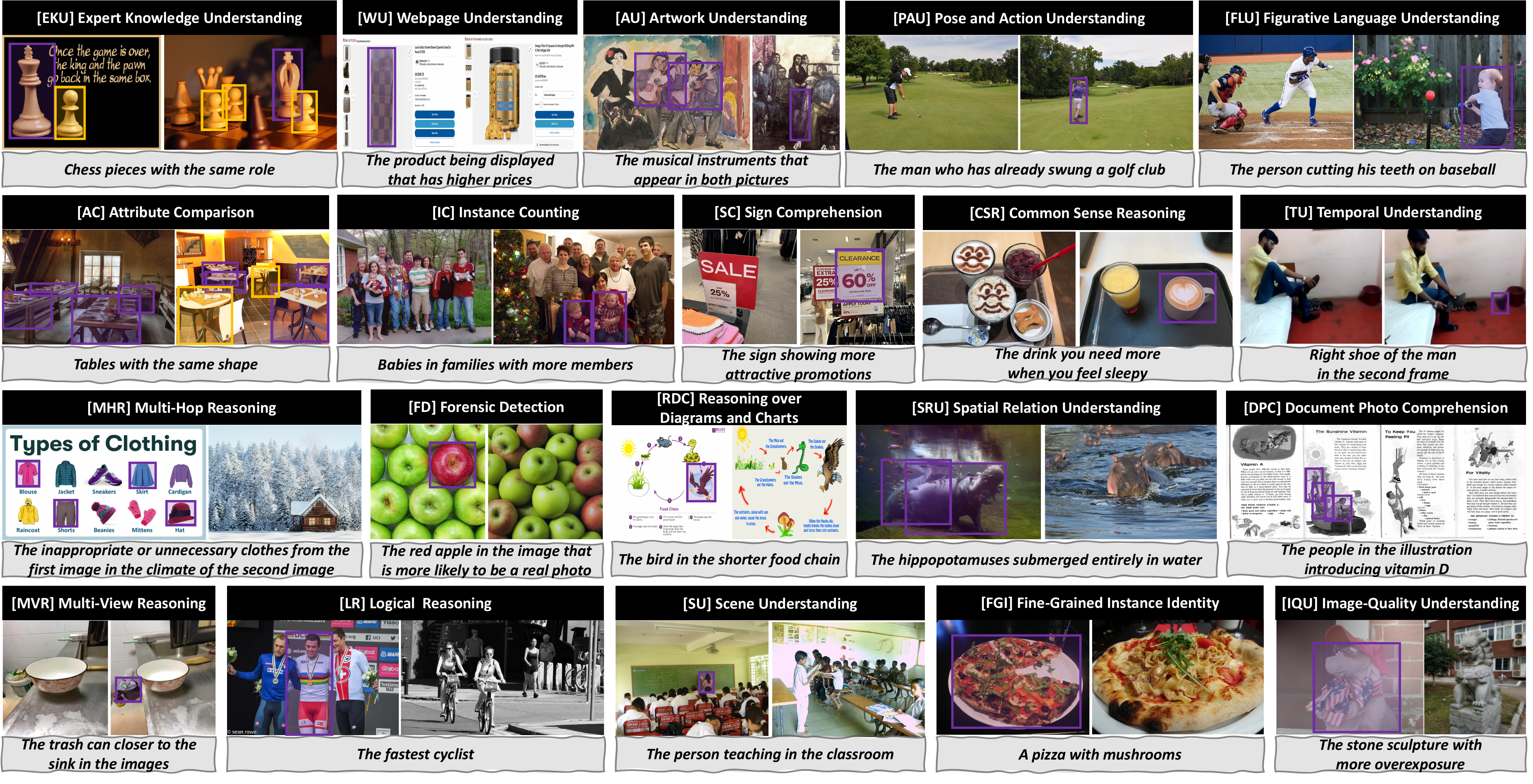}
\vspace{-2.5em}
\caption{MC-Bench contains diverse samples covering 20 practical skills.}
\label{fig: task examples}
\vspace{-1em}
\end{figure*}

Grounding visual content guided by textual inputs is a long-standing research topic involving vision-language understanding and visual localization tasks.
Early works typically focus on locating instances of interest using simple textual expressions, such as object detection (OD)~\cite{ren2015faster,xu2021pyramidal,redmon2016you,xu2022h2fa,carion2020end} and open-vocabulary object detection (OVD)~\cite{dhamija2020overlooked,joseph2021towards} based on category names, as well as referring expression comprehension (REC)~\cite{subramanian2022reclip,han2023zero,wang2024balanced,li2023transformer,li2021proposal} and describe object detection (DOD)~\cite{NEURIPS2023_f9fd24fd,schulter2023omnilabel} with referring phrases.
However, text descriptions in real-world applications are often more flexible and ambiguous. 
Grounding objects using free-form textual descriptions in an open world is challenging, as models must comprehend the intentions of ambiguous text inputs and grasp the overall context within the images.
Recently, the development of foundation models~\cite{kirillov2023segment,liu2023grounding,APE,xiao2024florence,ravi2024sam} has catalyzed a shift from specialized models to general-purpose models, showcasing unprecedented generalization capabilities.
Despite significant progress made by these foundation models, they still often struggle with complex text descriptions, limiting their broader applications for real-world use.

Since the advent of multimodal large language models (MLLMs)~\cite{awadalla2023openflamingo,achiam2023gpt,NEURIPS2023_9a6a435e,liu2024visual,li2023blip,zhu2023minigpt,wang2024protchatgpt,gao2023llama,zhang2025flexselect,xu2024gg}, these models have advanced significantly, demonstrating extraordinary capabilities in understanding human language and reasoning about the visual world.
Besides solving image-level visual-language tasks such as captioning~\cite{awadalla2023openflamingo,jia2024mos2} and visual question answering (VQA)~\cite{antol2015vqa,chen2025mathflow}, some recent MLLM works~\cite{chen2023shikra,zhang2024ferret,zhan2025griffonv1,chen2024revisiting,cai2025naver} have also explored more fine-grained tasks, showcasing promising region understanding and visual grounding capabilities. 
Despite their significance, we notice that, like many early visual grounding works, previous region-level MLLMs typically focus on single-image inputs, ignoring the cross-image context.

We believe that multi-image vision-language intelligence plays a pivotal role in many real-world applications, where the ability to extract and integrate contextual information from multiple images provides essential cues that enhance complex comprehension and reasoning. 
For instance, in autonomous driving, models~\cite{shao2024lmdrive,ding2024holistic} can better understand pedestrians and vehicles in the 3D world by integrating data from multiple camera angles.
In security and surveillance, models~\cite{chen2024videollm,ren2024timechat} can enhance system understanding of the dynamic environment by integrating multiple frames from different cameras to identify and analyze the targets across different time and locations.
General-purpose AI assistants (\eg, chart analysis~\cite{zhu2024multichartqa} and GUI agents~\cite{yan2023gpt}) are capable of understanding and reasoning across multiple contexts to identify correlations/discrepancies and make decisions.
Although some early works investigate vision-language intelligence in multi-image scenarios, they are limited to image-level tasks~\cite{tanaka2023slidevqa,liu2024mibench} or without complex textual descriptions~\cite{tang2014co,jiang2019class}.

Driven by this intuition, this paper explores a significant yet largely overlooked scenario and introduces a practical multi-image instance-level task, namely multi-context visual grounding, to assess such unproven abilities of existing MLLMs. 
This new task focuses on reasoning and localizing regions of interest across multiple images based on open-ended text prompts. 
As illustrated in Figure~\ref{fig: task compare}, compared to existing language-based visual grounding tasks~\cite{subramanian2022reclip,han2023zero,wang2024balanced,NEURIPS2023_f9fd24fd,qu2024rio,schulter2023omnilabel,lu2024zero}, multi-context visual grounding is more challenging, as it takes cross-image context into consideration and uses more nuanced and flexible textual expressions along with a greater diversity of disciplines.

To facilitate the research, we present MC-Bench, the first MLLM benchmark specifically designed for visual grounding in multi-image scenarios.
MC-Bench comprises 2,000 manually labeled samples, each featuring paired images, instance-level annotations and a corresponding text prompt.
The text prompts are categorized into three distinct styles (\ie, referring, comparison and reasoning), covering 20 practical skills applicable to real-world scenarios (see Figure~\ref{fig: task examples}).
Overall, we collect 3,345 diverse images from over 10 data sources, covering natural images, charts, document photos, artworks and scientific diagrams.
We then carefully curate 2,000 image pairs and manually annotated 1,514 unique open-ended text prompts, along with 3,200 language-grounded bounding boxes.

\begin{table*}[t!]
\footnotesize
\caption{Comparison to related vision-language datasets from different dimensions, \ie, multi-image input, instance-level annotation, multi-domain data and text description types. \hcmark~in the multi-image column indicates datasets containing multi-image subsets.}
\vspace{-1em}
\label{tab: dataset compare}
\centering
\setlength{\tabcolsep}{9pt}
\begin{tabular}{lcccc}
\toprule
Datasets & multi-image & instance-labeled & multi-domain & text description types \\
\midrule
MS-COCO~\cite{lin2014microsoft} & \xmark & \cmark & \xmark &  object categories \& image-level captions \\
RefCOCO/g/+~\cite{kazemzadeh2014referitgame,mao2016generation} & \xmark & \cmark & \xmark & category/attribute/relation descriptions \\
RIO~\cite{qu2024rio} & \xmark & \cmark & \xmark &  sentences of intention descriptions for objects\\
D$^3$~\cite{NEURIPS2023_f9fd24fd} & \xmark & \cmark & \cmark & unrestricted descriptions for any number of instances \\
OmniLabel~\cite{schulter2023omnilabel} & \xmark & \cmark & \cmark & complex object descriptions for any number of instances\\
ODinW~\cite{NEURIPS2022_3c4688b6} & \xmark & \cmark & \cmark  & object categories \& external knowledge descriptions \\
VQS~\cite{gan2017vqs} & \xmark & \cmark & \xmark & multi-choice QAs from the VQA dataset~\cite{antol2015vqa} \\
VizWiz-VQA-G~\cite{chen2022grounding} & \xmark & \cmark & \xmark & multi-choice QAs from the VizWiz-VQA dataset~\cite{gurari2018vizwiz} \\
\midrule
MMBench~\cite{liu2023mmbench} & \hcmark & \xmark & \cmark & multiple-choice QAs covering multiple ability dimensions \\
MMMU~\cite{yue2023mmmu} & \hcmark & \xmark & \cmark & multi-choice \& open QAs covering diverse disciplines \\
SEED-Bench~\cite{li2024seed} & \hcmark & \xmark & \cmark & multi-choice QAs spanning numerous dimensions\\
BLINK~\cite{fu2024blink} & \hcmark & \xmark & \cmark & multi-choice QAs on visual perception abilities \\
MileBench~\cite{song2024milebench} & \cmark & \xmark & \cmark & multi-choice \& open QAs on long video \& image sequences \\
Mantis-Eval~\cite{jiang2024mantis} & \cmark & \xmark & \cmark & multiple-choice \& open QAs on image sequences \\
MICBench~\cite{wu2024towards} & \cmark & \xmark & \cmark & multi-choice QAs on comparing image quality \\
Mementos~\cite{wang2024mementos} & \cmark & \xmark & \cmark & descriptions capturing unfolding events on image sequences \\
\midrule
MC-Bench (ours) & \cmark & \cmark & \cmark  & open-ended instance-level descriptions over multiple images \\
\bottomrule
\end{tabular}
\vspace{-1em}
\end{table*}

We evaluate over 20 baselines with potential multi-context visual grounding capabilities on MC-Bench, including advanced MLLMs and a few relevant foundation models without LLMs.
The experimental results indicate that current MLLMs have significant potential for improvement.
Concretely, while small-scale MLLMs (no larger than 7B) can achieve comparable instance-level performance to the foundation models~\cite{liu2023grounding,APE}, they typically show better image-level performance.
As MLLMs scale up, their performance improves significantly on all metrics.
We also observe that the specialist MLLMs trained exclusively on single-image visual grounding data struggle with multi-context scenarios.
In contrast, some generalist MLLMs with strong instruction-following capabilities generalize better in multi-context visual grounding, particularly those trained with multi-context data, even if those data are not instance-level labeled.
Nevertheless, a simple agentic baseline that integrates the strengths of GPT-4o~\cite{achiam2023gpt} and G-DINO~\cite{liu2023grounding} can easily outperform all evaluated end-to-end MLLMs by a clear margin, highlighting the potential for improvement.
We also introduce a fine-tuned baseline that is trained using synthesized multi-context instruction tuning data.
Moreover, we conduct human evaluations to establish an upper bound for existing MLLMs, revealing a significant performance gap between MLLMs and humans.

We hope our MC-Bench and empirical findings inspire the research community to delve deeper to discover and enhance the untapped potentials of MLLMs in instance-level tasks particularly in multi-image scenarios.
The main contributions of this paper can be summarized as follows:
\begin{itemize}
\item To the best of our knowledge, this work is the pioneer to explore the use of MLLMs for multi-image instance-level scenarios in open environments, and suggests a practical multi-context visual grounding task. 
\item We construct a new dataset, MC-Bench, featuring 2,000 manually annotated samples consisting of image pairs, text prompts, and corresponding instance-level labels. The diverse images and the open-ended prompts enable the evaluation of MLLMs from a wide range of dimensions.
\item We benchmark more than 20 relevant MLLMs and foundation models on MC-Bench, revealing a non-trivial performance gap between existing MLLMs and humans. Beyond the performance scores, this work provides insightful analysis aimed at guiding improvements in MLLM development.
\end{itemize}

\begin{figure*}[t]
\centering
\includegraphics[width=\linewidth]{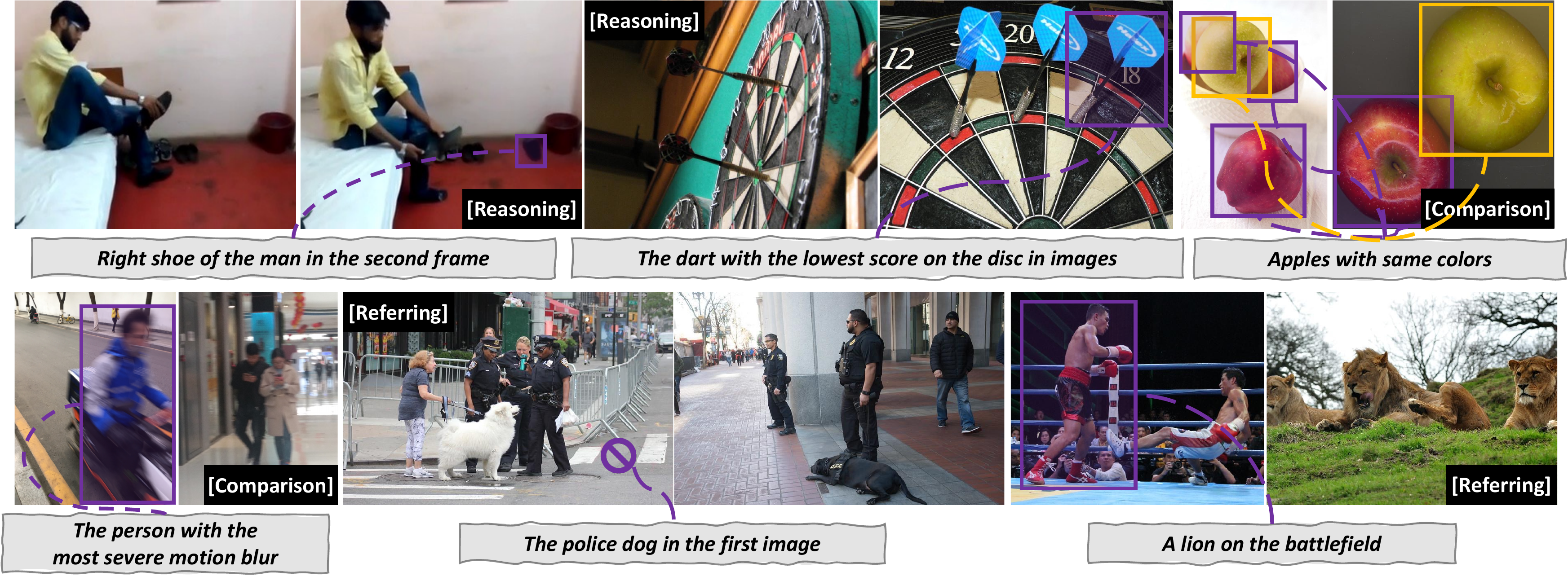}
\vspace{-2.5em}
\caption{MC-Bench contains three distinct styles of open-ended textual descriptions, \ie, referring, comparison and reasoning.}
\label{fig: features of tasks}
\vspace{-1em}
\end{figure*}

\section{Related Work}
\noindent\textbf{MLLM Benchmarks.}
Numerous benchmarks evaluate MLLMs with single-image inputs, and the assessments of the multimodal capabilities with multiple images do not receive much attention. 
Only a few recent benchmarks take multi-image evaluations into consideration, where some of them focus on specific domains and tasks (\eg, low-level vision~\cite{wu2024qbench,wu2024towards,wu2025bvinet} and temporal understanding~\cite{li2023empowering,li2023mvbench}).
As summarized in Table~\ref{tab: dataset compare}, some concurrent works~\cite{yue2023mmmu,song2024milebench,liu2023mmbench,jiang2024mantis,fu2024blink} present multi-image MLLMs benchmarks for more general purposes, covering multiple fields and disciplines.
However, they are annotated for image-level perception, comprehend and reasoning tasks (\eg, VQA), none of them is designed for instance-level tasks.
Current MLLMs for instance-level tasks are usually evaluated on conventional benchmarks~\cite{chen2022grounding,gan2017vqs,hudson2019gqa,kazemzadeh2014referitgame,mao2016generation,chen2024revisiting} with limited diversity and no multi-image context.

\vspace{0.3em}\noindent\textbf{Open-Ended Visual Content Grounding.}
Benefiting from the pre-trained visual-language models~\cite{radford2021learning,NEURIPS2022_ea370419}, open-vocabulary object detection~\cite{xu2023cyclic,gu2021open,zareian2021open,minderer2022simple} has received increasing attention, which localizes objects of arbitrary categories using language to achieve zero-shot transferability.
Besides leveraging category names, another line of work~\cite{subramanian2022reclip,han2023zero,wang2024balanced,NEURIPS2023_f9fd24fd,qu2024rio} investigates grounding visual content using simple referring phrases or sentences that often include auxiliary cues that help distinguish specific instances from others within the same category.
With the impressive success of LLMs, MLLMs have emerged as a pivotal advancement that serves to effectively connect vision and language tasks.
While MLLMs~\cite{NEURIPS2023_9a6a435e,liu2024visual,achiam2023gpt,li2023blip,awadalla2023openflamingo,zhu2023minigpt,gao2023llama} demonstrate remarkable capabilities on image-level tasks, several recent studies~\cite{lin2023sphinx,you2023ferret,chen2023shikra,zhao2023bubogpt,rasheed2023glamm,zhang2024groundhog,ma2024groma,zhang2023llava,guo2024regiongpt,wang2023cogvlm,pi2023detgpt,jiao2025lumen} explore the potential of enabling MLLMs to perform region-level tasks through instruction tuning.
However, most of existing works only focus on independent images and ignore multi-image context.

\vspace{0.3em}\noindent\textbf{MLLMs with Multi-Image Context.}
Unlike most previous MLLLMs take single-image-text pairs as inputs, some variants of MLLMs~\cite{zhang2023video,luo2023valley,ma2023vista} tailored for video tasks inherently support multiple frames and long contexts. 
However, these models designed to comprehend temporal sequences often face challenges when dealing with single images or multiple images that are not related temporally.
Another line of work~\cite{alayrac2022flamingo,awadalla2023openflamingo,li2023empowering,chen2023minigpt,bai2023qwen,li2024llava,laurenccon2025matters} has also noticed the importance of multiple-image capabilities for real-world applications, and takes effort for scaling the context to enable MLLMs to handle multiple and interleaved image-text inputs. 
Nevertheless, prior MLLMs largely neglect the multi-image instance-level scenarios, except for a few co-current works~\cite{li2025migician,nguyen2025calico,wahed2024prima} exploring common/unique objects/parts co-localization or simple co-referring.

\section{MC-Bench}
\subsection{Multi-Context Visual Grounding}\label{sec: task definition}
\noindent\textbf{Visual Grounding with Multi-Image Context.}
To meet the demands of open-ended real-world applications, this paper suggests a practical multi-image, instance-level vision-language task called multi-context visual grounding.
Given a multimodal input sample, \ie, multiple images and a text prompt, the models are required to localize all instances referenced in the input text description.
Each image in an input sample is temporally, spatially or semantically related with others, with the text prompt linking them through shared concepts or relationships.
Without loss of generality, we initially set the number of multi-images in the input samples to a pair, which maintains essential characteristics of multi-image tasks while ensuring a clear and controlled evaluation.
Our evaluation pipeline and metrics can be seamlessly extended to more challenging long-context scenarios.

\vspace{0.3em}\noindent\textbf{Visual Grounding with Open-Ended Expressions.}
Multi-context visual grounding aims at localizing specific instances within images using flexible and diverse text prompts, covering a broad range of practical skills.
As illustrated in Figure~\ref{fig: features of tasks}, we design three distinct styles of text prompts for grounding: referring, comparison and reasoning.
The referring style prompts identify instances using their category, attributes or positional information, either directly or indirectly.
The comparison style prompts are slightly more challenging, requiring models to ground instances by comparing the visual content across multiple images.
These comparisons can be global, based on image-level cues (\eg, the quantity of objects and image quality), or local, focusing on the attributes (\eg, color and shape) of objects within the images.
The reasoning style prompts describe instances in a more challenging manner, where models struggle to locate instances without relying on external knowledge (\eg, common sense and multi-hop reasoning skills) beyond the input itself.

\vspace{0.3em}\noindent\textbf{Visual Grounding with One-to-Any Matching.}
Since the text descriptions in multi-context visual grounding are unrestricted, each positive sample includes a text prompt that may refer to one or multiple instances within the images of that sample. 
In contrast, the text prompts in negative samples describe no instance within the images, and the models are encouraged to reject these negative inputs (\eg, the police dog example in Figure~\ref{fig: features of tasks}).
Textual expressions in the real world often exhibit high generalization and polysemy.
Therefore, we also assume that the models can accurately understand the intent behind flexible prompts and group target instances accordingly.
As shown in the top right of Figure~\ref{fig: features of tasks}, given images featuring apples of two colors and a prompt \textit{`Apples of the same colors'}, the models are encouraged not only to detect all the apples but also to group them according to their colors.

\subsection{Dataset Curation}\label{sec: data collect}
To the best of our knowledge, there is no existing dataset suitable for language-grounded cross-image instance-level tasks like multi-context visual grounding. 
To facilitate the research, we construct an evaluation-only dataset that effectively and faithfully evaluates the multimodal comprehend, reasoning and grounding capabilities of existing MLLMs in multi-image scenarios.

\vspace{0.3em}\noindent\textbf{Multi-Source Image Collection.}
Our goal is to create a high diverse benchmark that can better simulate a variety of real-world scenarios.
Guided by such goal, we first select images covering a wide range of domains and topics, \eg, natural images, comics, scientific diagrams, artworks, document photos, webpage screenshots, synthesized images and \etc.
Unlike conventional benchmarks, we emphasize instance-level tasks in real-world scenarios and collect a more extensive set of scene-centric images featuring a variety of object sizes and domains.
In total, we incorporate images from multiple data sources, including more than 10 existing datasets~\cite{lin2014microsoft,wu2023advancing,wu2024qbench,jiang2024mantis,mathew2021docvqa,fu2024blink,park2019robust,wu2024star,suhr2019corpus,NEURIPS2022_a96fe863,li2024seed2plus} and a few additional images crawled from the Internet. Please refer to the Appendix for more details.

\vspace{0.3em}\noindent\textbf{Linking Images through Text Descriptions.}
We then repurpose the collected images and link image pairs using open-ended text descriptions.
Concretely, the images are grouped into distinct subsets based on similar themes or domains.
The annotators are tasked with selecting image pairs from the subsets and writing an open-ended text prompt for each selected image pair, where the text prompts are supposed to properly leverage the cross-image context and clearly identify instances.
In addition, to facilitate the subsequent annotation process, annotators are asked to assign positive/negative flags to indicate whether the images contain at least one instance described by the text prompt.

\vspace{0.3em}\noindent\textbf{Instance-Level Labeling and Cyclic Review.}
After labeling the text descriptions for each image pair, we distribute the triplets to other annotators for subsequent annotation.
Given textual descriptions written by the text annotators, the box annotators are tasked with identifying the relevant instances within the positive samples and drawing bounding boxes to enclose them.
Once all the samples have instance-level annotations, we reassign them to the annotators who label the text prompts, asking them to review the bounding boxes to ensure they properly encompass the target instances indicated by the written prompt.
If any inconsistencies are found, the samples will be flagged for relabeling as part of the quality control process.
We build an online annotation platform based on Label Studio~\cite{LabelStudio}, leveraging its programmable and user-friendly interface for annotating paired images (see the interface example in the Appendix).

\begin{figure}[t]
\centering
\begin{subfigure}[t]{0.49\linewidth}
\centering
\includegraphics[width=0.75\linewidth]{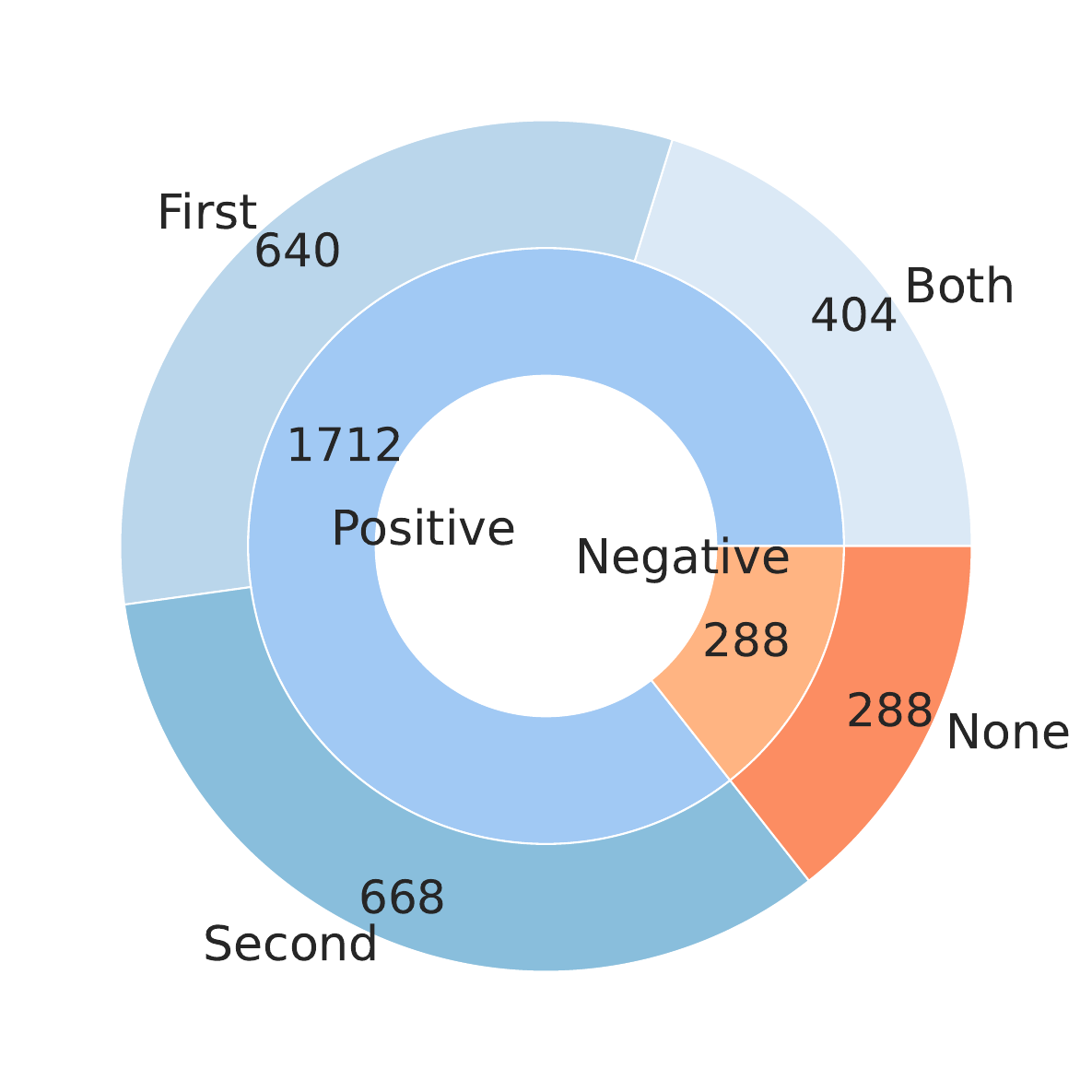}
\vspace{-0.9em}
\caption{distribution of labeled samples}
\label{fig: positive negative}
\end{subfigure}
\begin{subfigure}[t]{0.49\linewidth}
\centering
\includegraphics[width=0.8\linewidth]{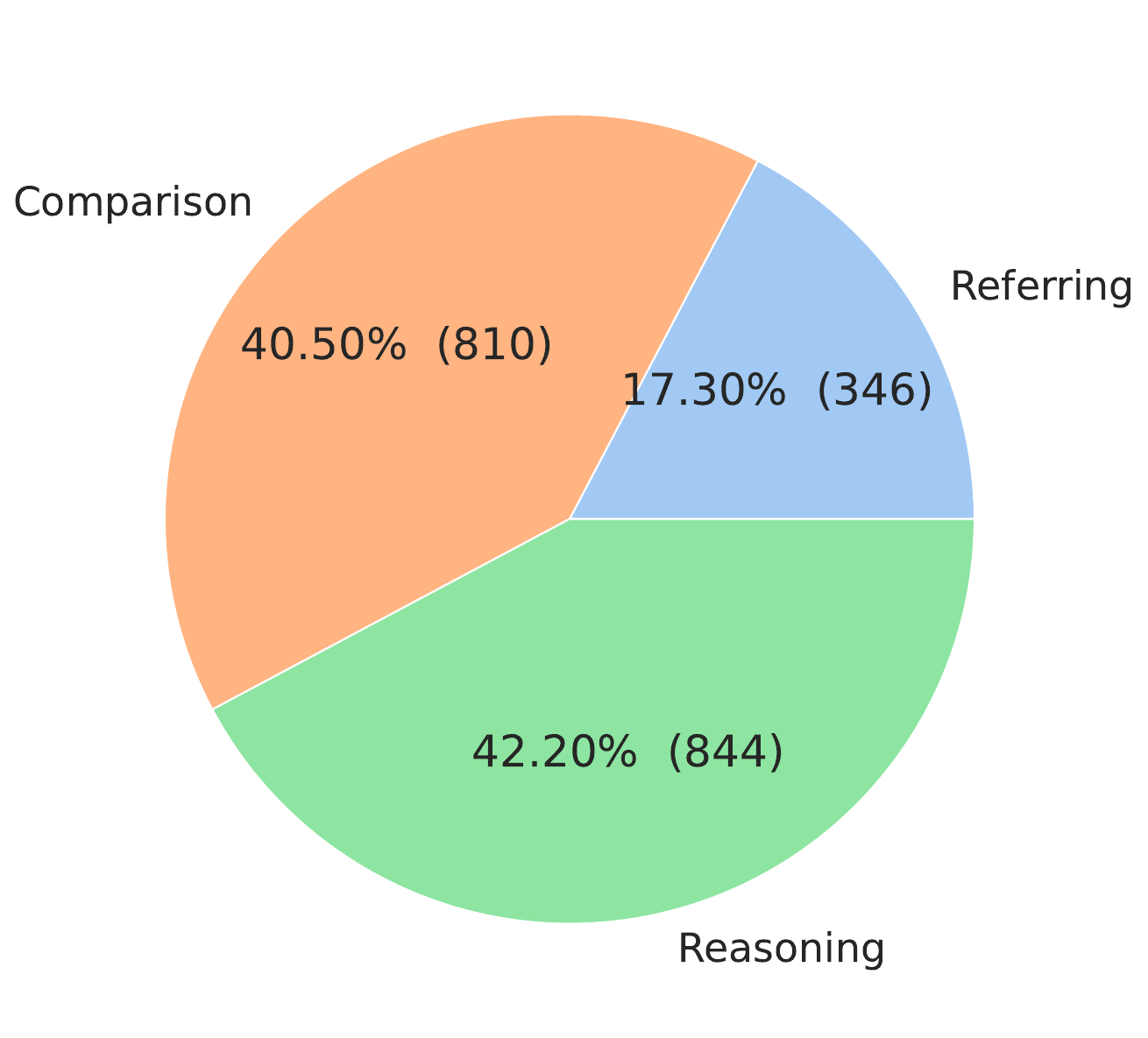}
\vspace{-0.9em}
\caption{ratio of text description types}
\label{fig: text type}
\end{subfigure}\\
\begin{subfigure}[t]{0.49\linewidth}
\centering
\includegraphics[width=\linewidth]{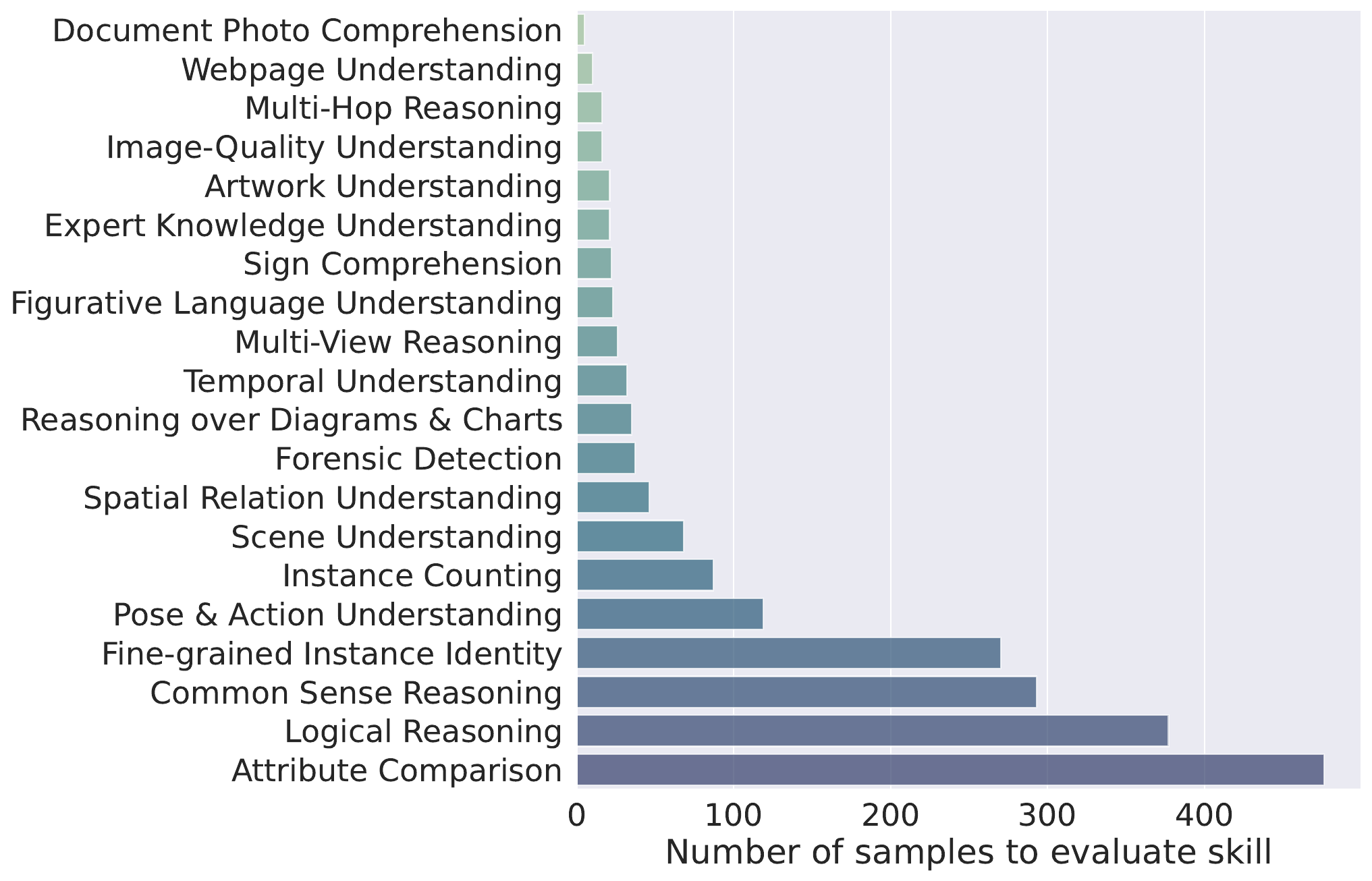}
\vspace{-1.5em}
\caption{distribution of covered skills}
\label{fig: skill type}
\end{subfigure}
\begin{subfigure}[t]{0.49\linewidth}
\centering
\includegraphics[width=\linewidth]{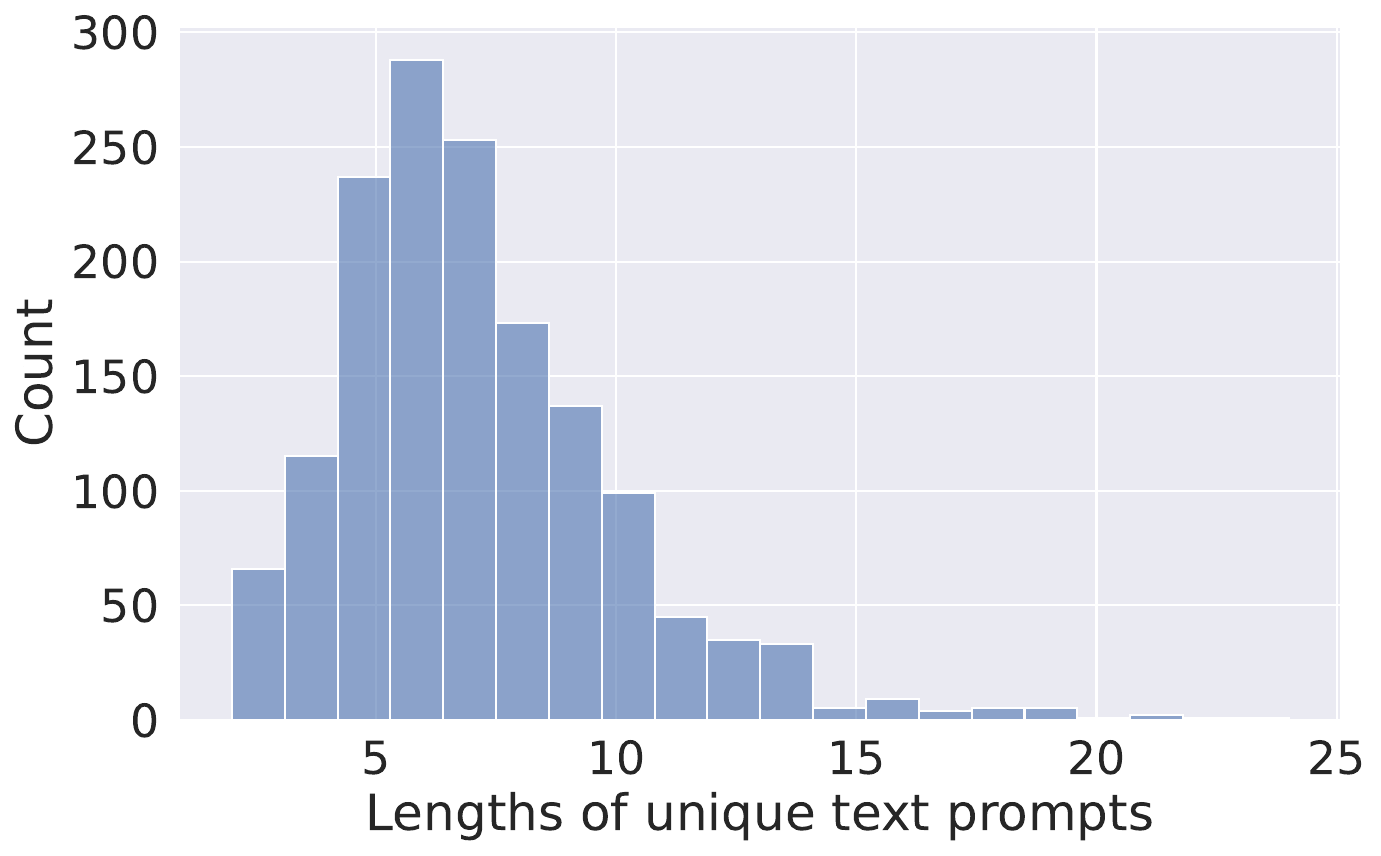}
\vspace{-1.5em}
\caption{distribution of text lengths}
\label{fig: text length}
\end{subfigure}\\
\begin{subfigure}[t]{0.49\linewidth}
\centering
\includegraphics[width=\linewidth]{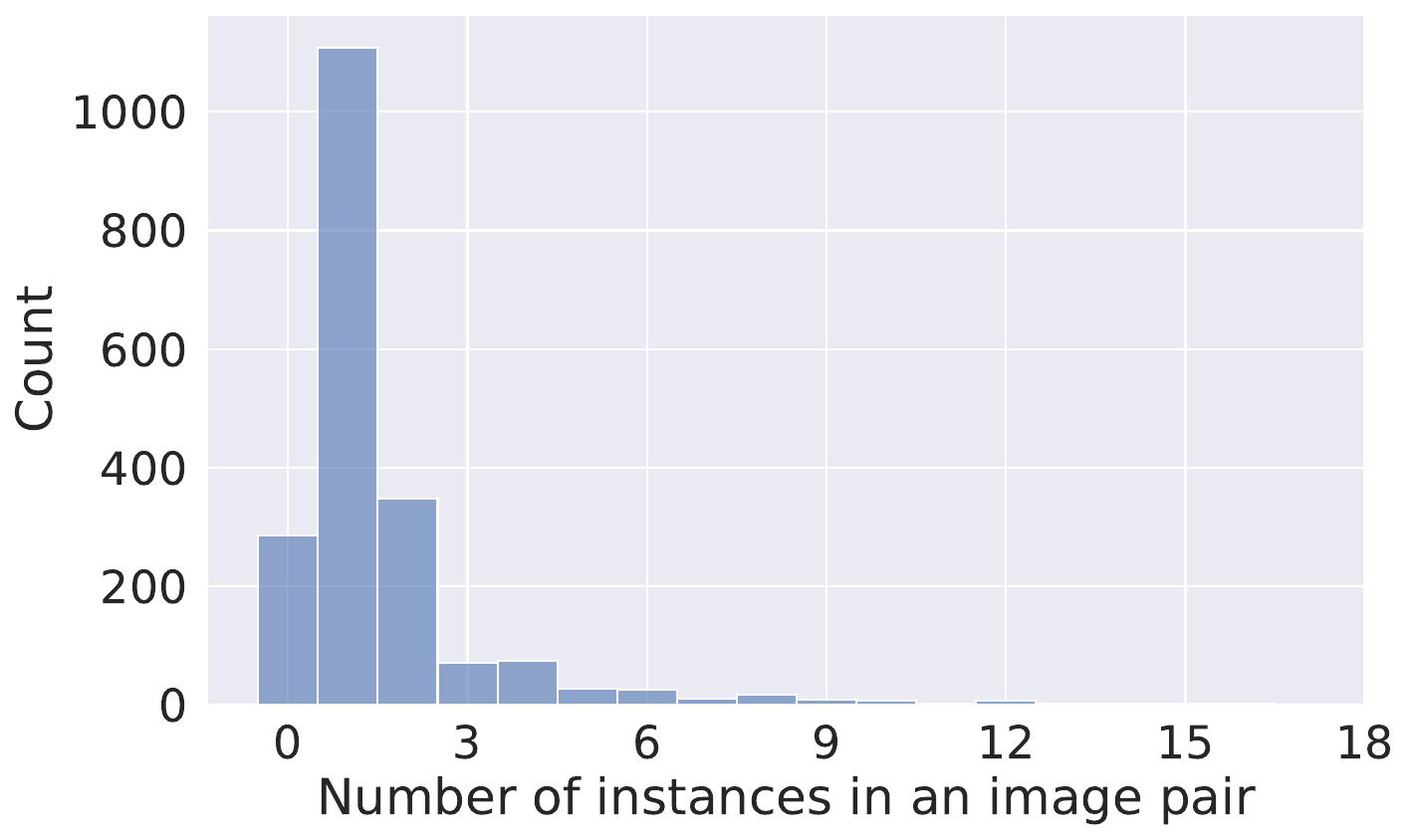}
\vspace{-1.5em}
\caption{distribution of instance quantity}
\label{fig: box num}
\end{subfigure}
\begin{subfigure}[t]{0.49\linewidth}
\centering
\includegraphics[width=\linewidth]{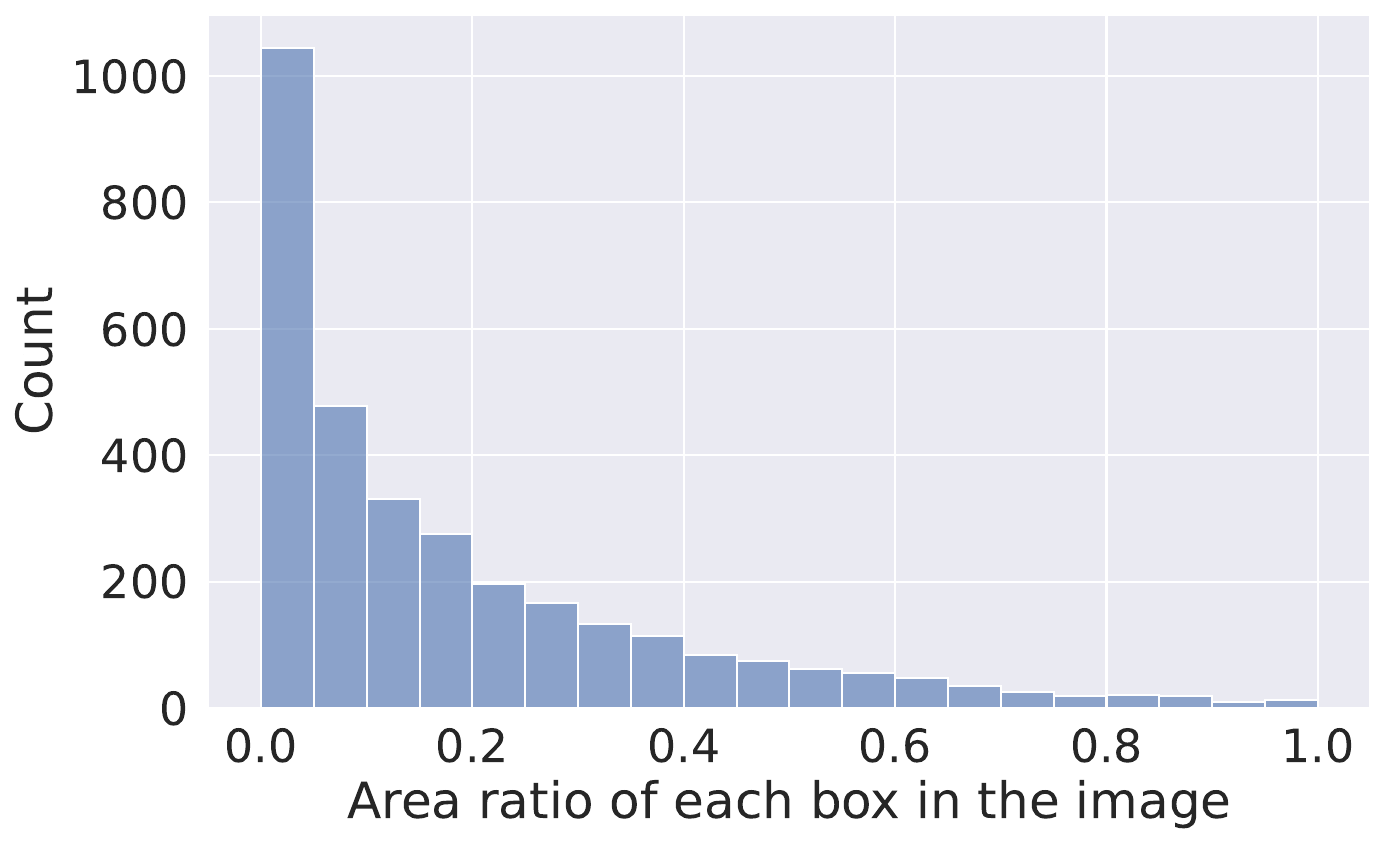}
\vspace{-1.5em}
\caption{distribution of instance sizes}
\label{fig: object size}
\end{subfigure}
\vspace{-1em}
\caption{Statistical analysis of the proposed MC-Bench.}
\label{fig: dataset statistics}
\vspace{-1em}
\end{figure}

\subsection{Dataset Statistics}\label{sec: dataset statistics}
We gather a total of 3,345 different images from various sources, covering various domains and topics.
We meticulously organize the collected images into 2,000 image pairs and provide 1,514 unique open-ended text descriptions for these image pairs. 
As shown in Figure~\ref{fig: text length}, the length of the text descriptions ranges from 2 to 24 words, with an average of 7.2.
Each text prompt describes visual content within paired images without restriction.
MC-Bench has 1,712 positive samples, with 404 containing target instances in both images, while the remaining samples having target objects in only one image (either the first or the second), as summarized in Figure~\ref{fig: positive negative}.
Besides positive samples, we add a small proportion of negative examples to evaluate the capabilities of models for rejecting negative inputs.
As illustrated in Figures~\ref{fig: text type} and \ref{fig: skill type}, MC-Bench contains three distinct styles of text expressions (\ie, 346, 810 and 844 for referring, comparison and reasoning respectively) and 20 practical skills (\eg, attribute comparison, logical reasoning, common sense reasoning and multi-view reasoning). 

For the instance-level annotations, MC-Bench includes 3,200 language-grounded bounding boxes in total.
As summarized in Figure~\ref{fig: box num}, each prompt in positive samples indicates 1 to 17 instances of 1 to 7 groups within image pairs, while there is no instance related to negative description.
Unlike benchmarks for image-level tasks, we collect more challenging scene-centric images and label instances with diverse sizes. 
The size of the labeled bounding boxes ranges from 4e-6 to 1, as shown in the distribution in Figure~\ref{fig: object size}.

\section{Experiments}
\subsection{Evaluation Metrics}\label{sec: metrics}
\noindent\textbf{Image-Level Metrics.}
For multi-context visual grounding task, we design image-level and instance-level metrics to evaluate the performance of models from different dimensions.
Accuracy (Acc) is used to confirm whether the models can correctly identify which images contain the objects indicated by each text prompt, where the instance quantity and fine-grained location information is not considered.

\vspace{0.3em}\noindent\textbf{Instance-Level Metrics.}
We choose average precision (AP$_{50}$) as the instance-level metric to verify whether the models can locate the target instances with multi-context inputs.
For samples where the text prompt describes multiple groups of instances, we first apply Hungarian algorithm to match each predicted group to the most appropriate ground-truth group, ensuring that the mean intersection over union (IoU) across all predictions is maximized.

\subsection{Baselines}
Since the multi-context visual grounding is a new task, we implement and evaluate various advanced approaches with potential visual grounding capabilities, including latest proprietary and open-source MLLMs as well as foundation models without LLMs. 
Most existing methods do not support multi-image inputs, and we horizontally concatenate the images before feeding them to these models.

Specifically, we select and evaluate 
\textbf{\ding{182} the API-based generalist MLLMs}, such as GPT-4o~\cite{achiam2023gpt} and Gemini-1.5 Pro~\cite{reid2024gemini}, 
\textbf{\ding{183} the open-source generalist MLLMs} (\eg, Qwen-VL series~\cite{bai2023qwen,Qwen2-VL}, SPHINX~\cite{lin2023sphinx}, InternVL2.5~\cite{chen2024expanding} and MiniGPT-v2~\cite{chen2023minigpt}) which are capable of performing a wide range vision-language tasks, 
\textbf{\ding{184} the open-source specialist MLLMs} (\eg, Shikra~\cite{chen2023shikra}, Kosmos-2~\cite{peng2023kosmos}, Ferret~\cite{you2023ferret}, Lenna~\cite{wei2023lenna}, Groma~\cite{ma2024groma} and GroundingGPT~\cite{li-etal-2024-groundinggpt}) tailored to visual grounding-related tasks and 
\textbf{\ding{185} the foundation models without LLMs}, such as ONE-PEACE~\cite{wang2023one}, G-DINO~\cite{liu2023grounding} and APE~\cite{APE}.
More details (\eg, model version and used prompts) are provided in the Appendix.

\begin{figure}[t]
\centering
\includegraphics[width=\linewidth]{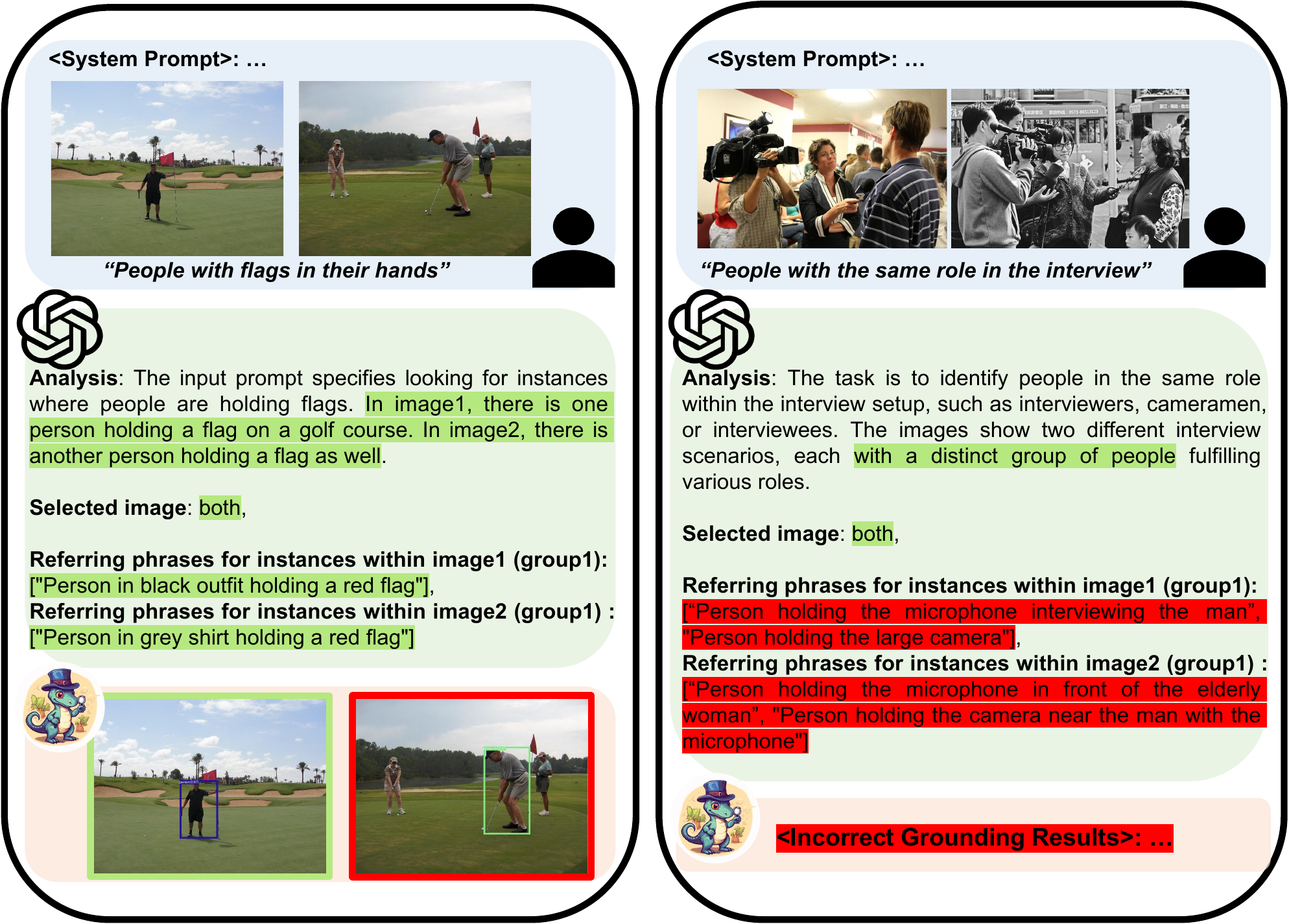}
\vspace{-2.25em}
\caption{Some case examples of the agentic baseline, where the correct and wrong predictions are highlighted using \textcolor{correctcolor}{green} and \textcolor{red}{red}. 
The left case shows the detection error caused by G-DINO, while the right case demonstrates the grouping error caused by GPT-4o.}
\label{fig: prediction example}
\vspace{-1em}
\end{figure}

Apart from aforementioned end-to-end approaches, we devise and assess \textbf{\ding{186} an agentic baseline} that follows a simple yet effective divide-and-conquer strategy~\cite{wei2022chain} and takes the advantages of MLLMs and detectors in reasoning and localization respectively.
Concretely, we utilize GPT-4o as a reasoning agent and prompt it to first analyze multi-context inputs to determine which images contain the target instances described by the text phrases.
This reasoning agent is then requested to generate concise and discriminative referring phrases for each individual target instance.
We finally localize the target objects using G-DINO~\cite{liu2023grounding} along with the GPT-generated referring phrases.
Some examples of our agentic baseline are showcased in Figure~\ref{fig: prediction example}.

We also introduce and evaluate \textbf{\ding{187} a finetuned baseline} that enhances existing end-to-end MLLM (\ie, Qwen2-VL-7B~\cite{Qwen2-VL}) by multi-context instruction tuning. 
We construct a multi-context instruction tuning dataset with over 50K samples by collecting multi-context image-level task samples from existing datasets~\cite{forbes2019neural,jiang2024mantis} and synthesizing multi-context instance-level task samples. 
To accelerate the training process/maintain the generalization capabilities of the MLLM, we finetune models with LoRA~\cite{hu2021lora}.
Please refer to the Appendix for more training details.

We conduct \textbf{\ding{188} human evaluations} to establish an upper bound for the models.
In total, we invite 3 volunteers who have not been exposed to annotated data to  participate in the evaluation with all 2K multi-context samples. 
Given each text prompt, the participants are asked to draw bounding boxes for the target instances in corresponding image pairs.

\begin{table*}[t]
\footnotesize
\caption{Comparison of baselines on MC-Bench. \textit{Sequence} indicates whether the model supports image sequences as inputs, where \hcmark\ denotes that some intermediate steps support image sequences. The superscripts \textit{ref}, \textit{com} and \textit{rea} denote the results for the three specific types respectively.}
\label{tab: performance compare}
\vspace{-1em}
\centering
\setlength{\tabcolsep}{7pt}
\begin{tabular}{lcccccccccc}
\toprule
   & & &\multicolumn{4}{c}{Image-Level} & \multicolumn{4}{c}{Instance-Level} \\
\cmidrule(r){4-7}\cmidrule(r){8-11}
Methods & sequence & LLM size & $\rm{Acc^{ref}}$ & $\rm{Acc^{com}}$ & $\rm{Acc^{rea}}$ & $\rm{Acc}$ & $\rm{AP_{50}^{ref}}$ & $\rm{AP_{50}^{com}}$ & $\rm{AP_{50}^{rea}}$ & $\rm{AP_{50}}$ \\
\midrule
\rowcolor{gray!15!white}\multicolumn{11}{c}{\textit{API-Based Generalist MLLMs}} \\
GPT-4o~\cite{achiam2023gpt} & \cmark & - & 69.7 & 82.8 & 77.5 & 78.3 & 1.8 & 3.9 & 2.3 & 2.8 \\
Gemini-1.5 Pro~\cite{reid2024gemini} & \cmark & - & 56.1 & 65.1 & 62.7 & 62.5 & 30.6 & 29.9 & 26.1 & 28.2 \\ 
\midrule
\rowcolor{gray!15!white}\multicolumn{11}{c}{\textit{Open-Source Generalist MLLMs}} \\
Qwen-VL-Chat~\cite{bai2023qwen} & \cmark & 7B & 33.8 & 34.8 & 31.8 & 33.4 & 10.9 & 9.2 & 9.0 & 9.3 \\ 
Qwen-VL-Chat~\cite{bai2023qwen} & \xmark & 7B & 36.7 & 47.7 & 45.5 & 44.9 & 21.7 & 17.3 & 17.0 & 17.5 \\
Qwen2-VL~\cite{Qwen2-VL} & \cmark & 7B & 43.9 & 60.1 & 54.3 & 54.9 & 22.5 & 21.3 & 16.2 & 19.1 \\ 
Qwen2-VL~\cite{Qwen2-VL} & \xmark & 7B & 43.6 & 52.2 & 53.7 & 51.4 & 19.9 & 18.0 & 17.5 & 17.8 \\
Qwen2-VL~\cite{Qwen2-VL} & \cmark & 72B & 61.6 & 79.1 & 68.0 & 71.4 & 33.7 & 33.2 & 27.0 & 30.7 \\ 
Qwen2-VL~\cite{Qwen2-VL} & \xmark & 72B & 43.1 & 53.5 & 52.8 & 51.4 & 29.6 & 26.7 & 24.4 & 26.0 \\
InternVL2.5~\cite{chen2024expanding} & \cmark & 8B & 28.0 & 37.7 & 38.5 & 36.4 & 15.7 & 10.9 & 9.6 & 11.1 \\ 
InternVL2.5~\cite{chen2024expanding} & \xmark & 8B & 38.7 & 54.8 & 53.2 & 51.4 & 12.9 & 11.4 & 10.1 & 10.8 \\
SPHINX-1k~\cite{lin2023sphinx} & \xmark & 13B & 41.9 & 49.6 & 51.1 & 48.9 & 16.2 & 15.8 & 14.0 & 14.9 \\ 
SPHINX-v2-1k~\cite{lin2023sphinx} & \xmark & 13B & 41.3 & 52.2 & 38.9 & 44.7 & 26.5 & 21.1 & 19.0 & 20.8 \\ 
MiniGPT-v2~\cite{chen2023minigpt} & \xmark & 7B & 34.1 & 43.8 & 45.6 & 42.9 & 11.7 & 12.2 & 10.8 & 11.6 \\ 
\midrule
\rowcolor{gray!15!white}\multicolumn{11}{c}{\textit{Open-Source Specialist MLLMs}} \\
Shikra~\cite{chen2023shikra} & \xmark & 7B & 37.6 & 44.7 & 45.4 & 43.8 & 10.0 & 10.6 & 9.1 & 9.8 \\
Kosmos-2~\cite{peng2023kosmos} & \xmark & 1.6B & 26.3 & 30.6 & 33.6 & 31.2 & 10.7 & 11.5 & 10.5 & 10.6  \\
Lenna~\cite{wei2023lenna} & \xmark & 7B & 30.3 & 30.6 & 28.6 & 29.7 & 17.1 & 14.3 & 12.7 & 13.9 \\
Groma~\cite{ma2024groma} & \xmark & 7B & 34.1 & 44.4 & 42.4 & 41.8 & 17.2 & 15.6 & 12.8 & 14.2 \\
GroundingGPT~\cite{li-etal-2024-groundinggpt} & \xmark & 7B & 35.5 & 43.3 & 46.3 & 43.3 & 14.4 & 12.2 & 11.9 & 12.3 \\
Ferret~\cite{you2023ferret} & \xmark & 7B & 34.4 & 42.6 & 45.5 & 42.4 & 12.8 & 12.6 & 9.5 & 11.0 \\
Ferret~\cite{you2023ferret} & \xmark & 13B & 35.8 & 44.7 & 48.6 & 44.8 & 13.4 & 13.5 & 12.3 & 12.9 \\
CogVLM-Grounding~\cite{wang2023cogvlm} & \xmark & 17B & 40.5 & 50.2 & 50.1 & 48.5 & 20.9 & 18.0 & 16.0 & 17.5 \\
\midrule
\rowcolor{gray!15!white}\multicolumn{11}{c}{\textit{Foundation Models without LLMs}} \\
G-DINO-B~\cite{liu2023grounding} & \xmark & \xmark & 31.2 & 30.4 & 30.9 & 30.8 & 13.9 & 15.6 & 15.3 & 15.0 \\
APE (D)~\cite{APE} & \xmark & \xmark & 24.0 & 20.6 & 16.2 & 19.3 & 20.4 & 20.8 & 16.1 & 18.8 \\
ONE-PEACE~\cite{wang2023one} & \xmark & \xmark & 32.9 & 42.7 & 42.3 & 40.9 & 17.8 & 15.5 & 10.2 & 13.3 \\
\midrule
Agentic Baseline$_\textit{GPT-4o+G-DINO}$ & \hcmark & - & 66.8 & 84.8 & 75.7 & 77.9 & 41.6 & 37.2 & 34.4 & 36.2 \\
Finetuned Baseline$_\textit{Qwen2-VL-7B}$ & \cmark & 7B & 47.1 & 59.9 & 60.0 & 57.7 & 26.7 & 23.2 & 20.8 & 22.6 \\
\midrule
\textcolor{gray}{Humans} & \textcolor{gray}{-} & \textcolor{gray}{-} & \textcolor{gray}{89.5} & \textcolor{gray}{95.4} & \textcolor{gray}{90.5} & \textcolor{gray}{92.3} & \textcolor{gray}{47.8} & \textcolor{gray}{40.3} & \textcolor{gray}{41.0} & \textcolor{gray}{41.3} \\
\bottomrule
\end{tabular}
\vspace{-1em}
\end{table*}

\subsection{Benchmark Results}
We divide existing approaches into different groups and report their performance in Table~\ref{tab: performance compare}.
The proprietary generalist MLLMs~\cite{liu2023grounding,reid2024gemini} are used through API calls and generally considered to have huge model sizes.
These models inherently support image sequence inputs and show strong image-level comprehend and reasoning capabilities.
However, while Gemini-1.5 Pro~\cite{reid2024gemini} achieves competitive instance-level performance, GPT-4o~\cite{achiam2023gpt} exhibits limited fine-grained localization capabilities.

For the open-source MLLMs accepting image sequence inputs (\ie, Qwen-VL-Chat~\cite{bai2023qwen}, Qwen2-VL~\cite{Qwen2-VL} and InternVL2.5~\cite{chen2024expanding}), we compare both sequence- and merge-image variants.
We find that as model capabilities increase (\ie, Qwen-VL to Qwen2-VL, and 7B to 72B LLM), the sequence-image variants more clearly exceed merge-image variants.
Among all tested open-source MLLMs~\cite{bai2023qwen,Qwen2-VL,chen2024expanding,lin2023sphinx,chen2023minigpt}, Qwen2-VL-72B with image sequence inputs achieves the best results, even outperforms proprietary MLLMs on instance-level metrics.

Generally, the specialist MLLMs~\cite{chen2023shikra,you2023ferret,li-etal-2024-groundinggpt,ma2024groma,wang2023cogvlm,wei2023lenna,peng2023kosmos} are specially designed or fine-tuned for visual grounding-related tasks. 
However, in multi-context visual grounding, existing specialists obtain worse results in terms of both image-level and instance-level metrics.
For instance, the largest specialist CogVLM-Grounding-17B~\cite{wang2023cogvlm} achieves performance comparable to some 7B generalist MLLMs (\eg, Qwen-VL-Chat and Qwen2-VL).
We attribute this to the limited generalization capabilities of these specialists tailored to single-image scenarios.

Compared to MLLM counterparts, the foundation models~\cite{liu2023grounding,APE,wang2023one} without LLMs still perform well on instance-level metrics.
However, these models tend to generate redundant low-confidence boxes within irrelevant images, leading to deteriorated Acc performance.
The agentic baseline integrates extraordinary multi-modal comprehension and reasoning capabilities of GPT-4o and excellent localization capabilities of G-DINO~\cite{liu2023grounding}, thereby achieving remarkable results and surpassing aforementioned end-to-end approaches.
We also observe that after multi-context instruction tuning, the cross-image perception and localization capabilities of Qwen2-VL-7B are significantly enhanced, leading to 2.8\% Acc and 3.5\% $\rm{AP_{50}}$ gains.
Moreover, we calculate the average results of all volunteers as the upper bound.
Human evaluations outperform the agentic baseline by 14.4\% in  $\rm{Acc}$ and 5.1\% in $\rm{AP_{50}}$, underscoring a clear performance gap between models and humans.

\begin{figure*}[t]
\centering
\begin{subfigure}[t]{0.33\linewidth}
\centering
\includegraphics[width=\linewidth]{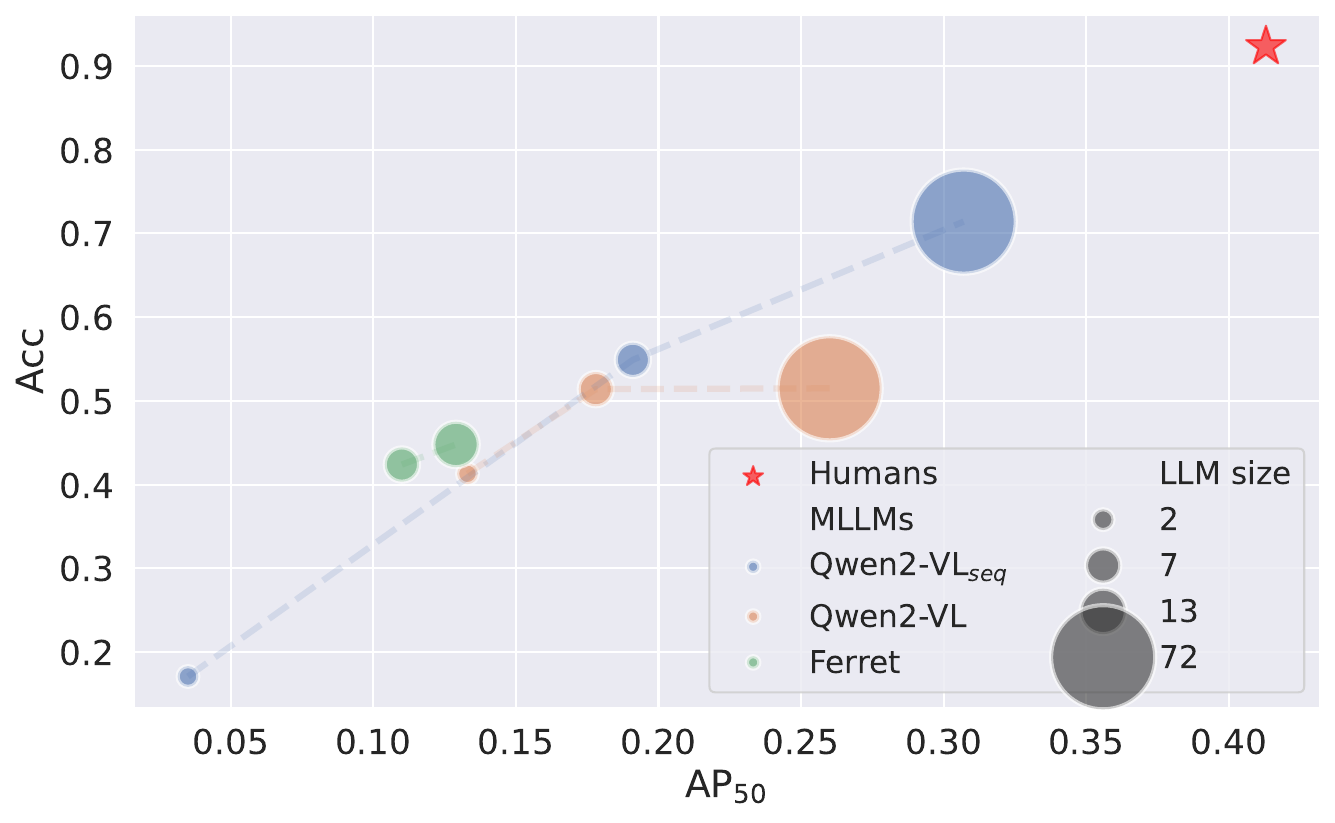}
\vspace{-1.6em}
\caption{effectiveness of model sizes}
\label{fig: model size analysis}
\end{subfigure}
\begin{subfigure}[t]{0.33\linewidth}
\centering
\includegraphics[width=\linewidth]{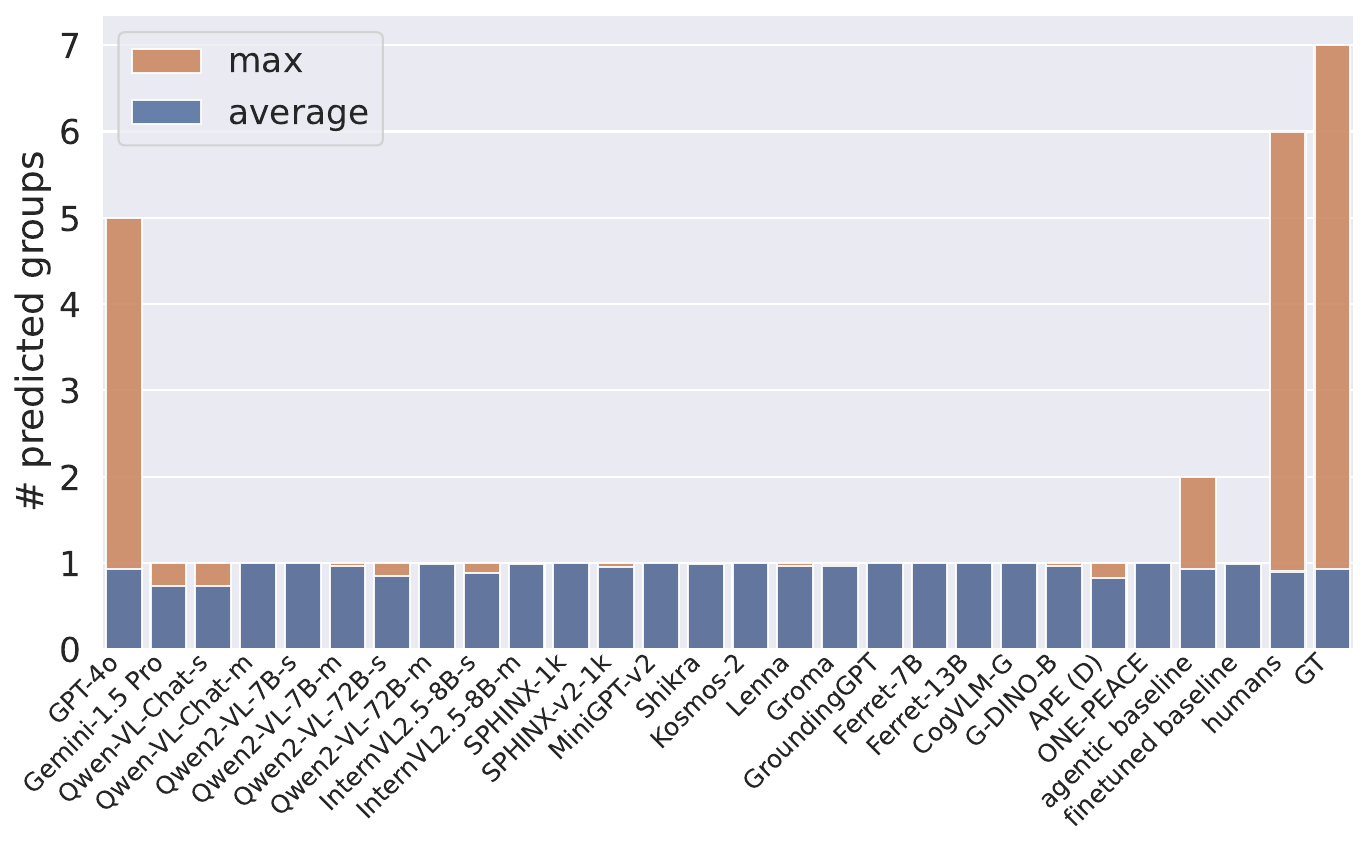}
\vspace{-1.6em}
\caption{number of predicted groups}
\label{fig: group analysis1}
\end{subfigure}
\begin{subfigure}[t]{0.33\linewidth}
\centering
\includegraphics[width=\linewidth]{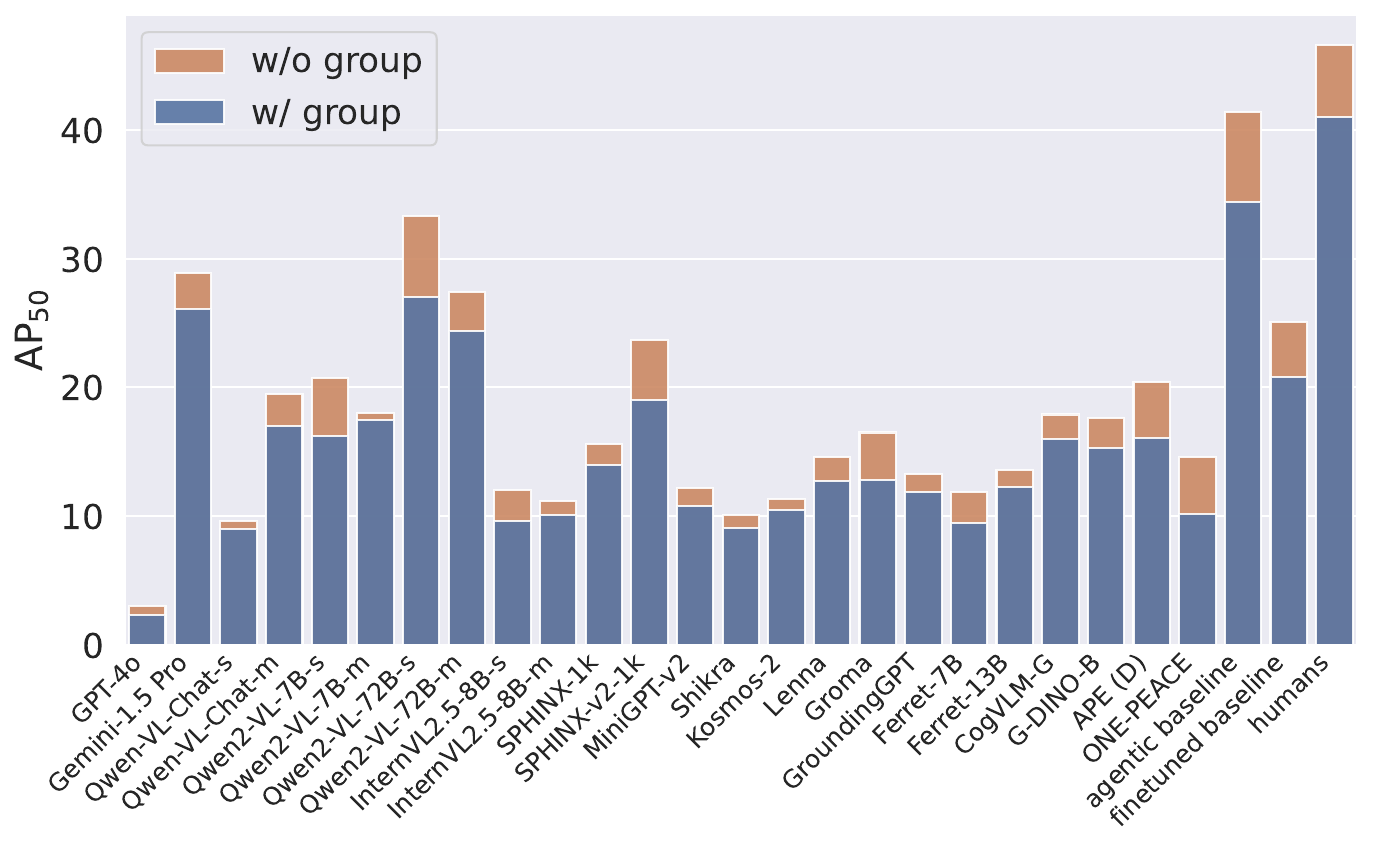}
\vspace{-1.6em}
\caption{group-agnostic performance}
\label{fig: group analysis2}
\end{subfigure}\\
\begin{subfigure}[t]{0.33\linewidth}
\centering
\includegraphics[width=\linewidth]{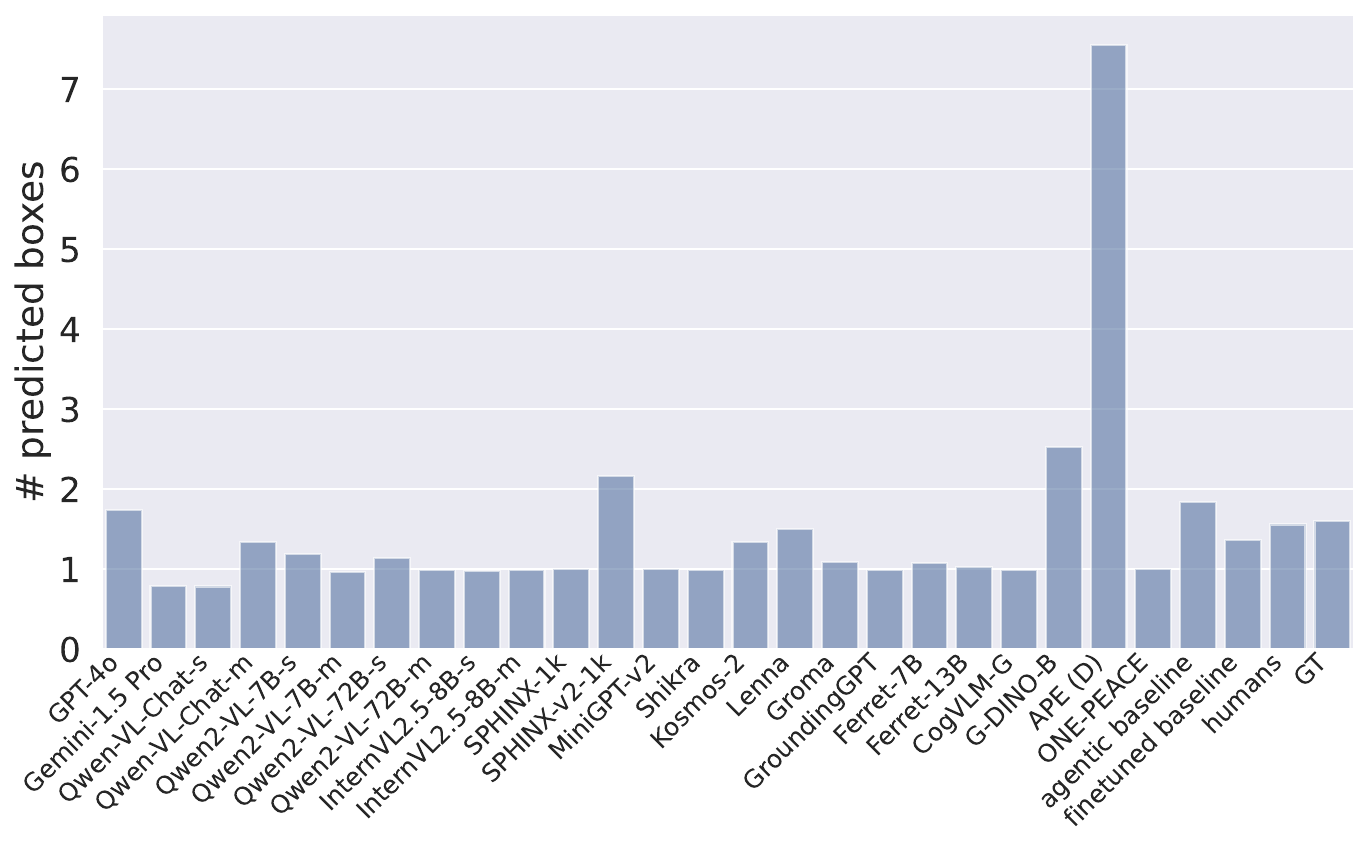}
\vspace{-1.6em}
\caption{average number of predicted boxes}
\label{fig: box analysis}
\end{subfigure}
\begin{subfigure}[t]{0.33\linewidth}
\centering
\includegraphics[width=\linewidth]{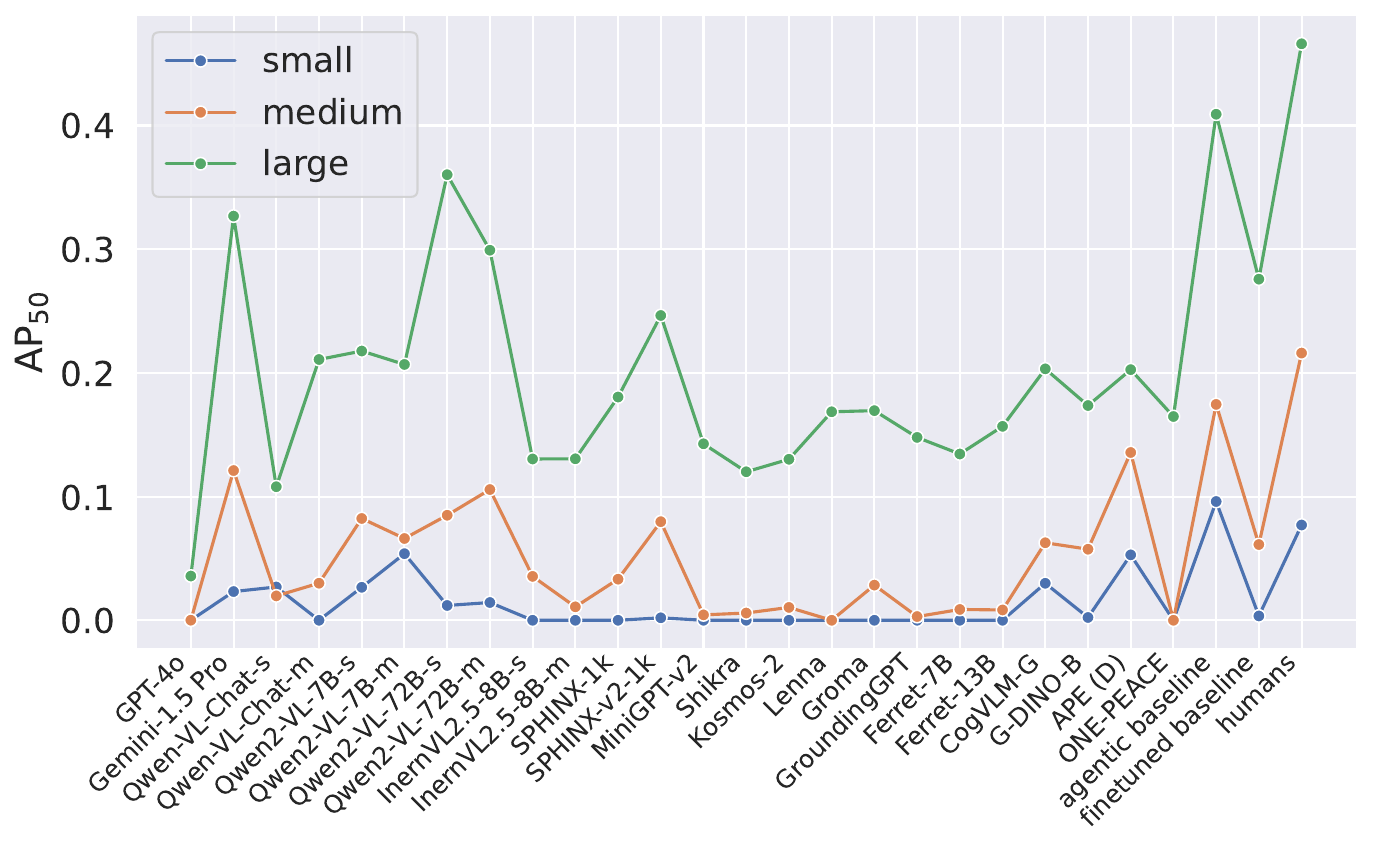}
\vspace{-1.6em}
\caption{results of different object sizes}
\label{fig: size analysis}
\end{subfigure}
\begin{subfigure}[t]{0.33\linewidth}
\centering
\includegraphics[width=\linewidth]{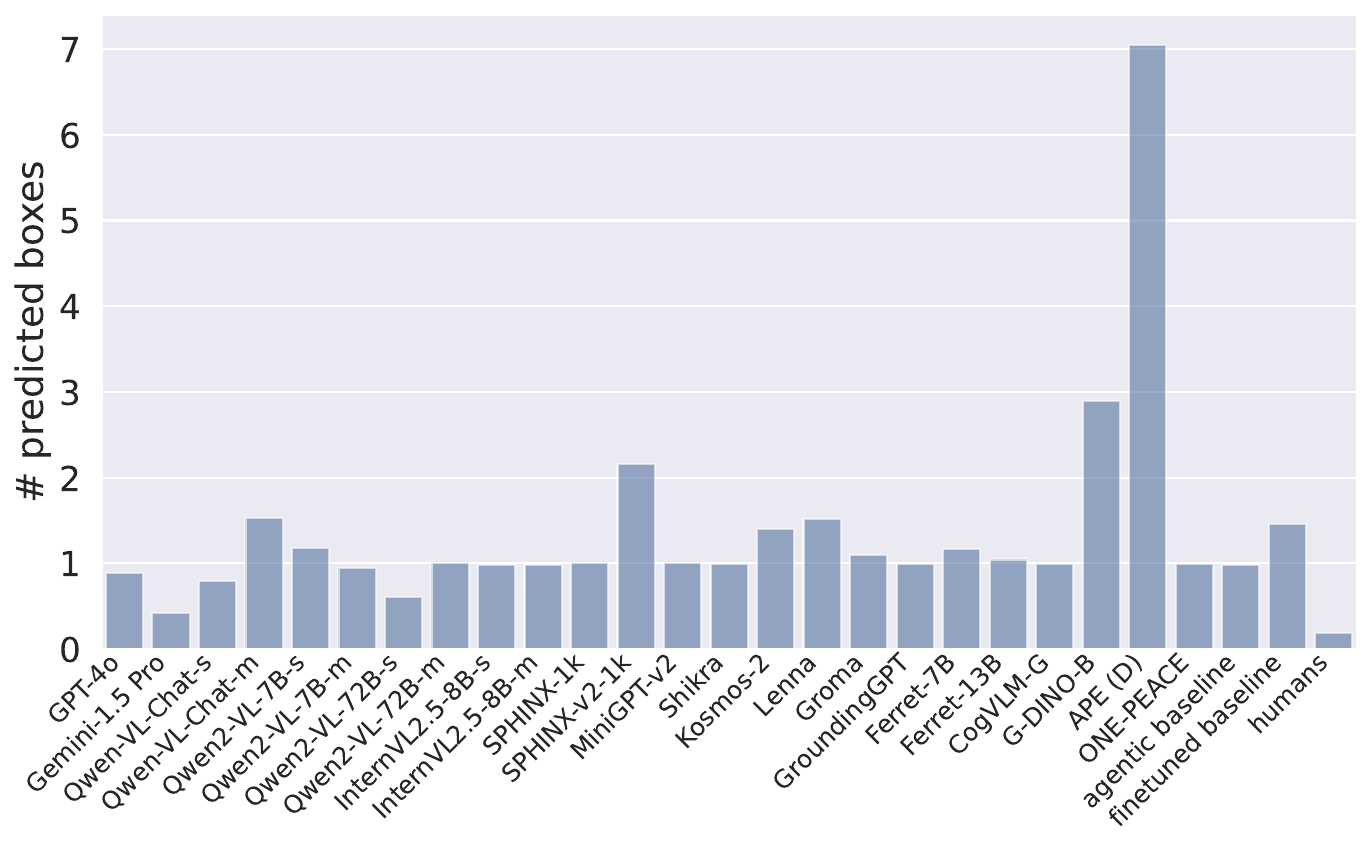}
\vspace{-1.6em}
\caption{average prediction count on negative samples}
\label{fig: negative analysis}
\end{subfigure}
\vspace{-1em}
\caption{More analysis experiments on MC-Bench.}
\vspace{-1em}
\end{figure*}

\subsection{More Analysis}
We conduct multiple analytical experiments to further explore MLLMs from different perspectives. 
For the open-source MLLMs (\eg, Qwen2-VL~\cite{Qwen2-VL} and Ferret~\cite{you2023ferret}) with various model size variants, we analyze the impact of model size, as visualized in Figure~\ref{fig: model size analysis}.
Larger models show sustained performance improvement on both Acc and AP$_{50}$, consistent with the scaling law~\cite{kaplan2020scaling}. 

In multi-context visual grounding, a single text prompt may describe objects from multiple groups. 
As shown in Figure~\ref{fig: group analysis1}, we observe that current approaches struggle with assigning groups, with most models predicting only one group.
By replacing the standard instance-level metric with a group-agnostic one, almost all baselines achieve significantly higher AP$_{50}$ results (see Figure~\ref{fig: group analysis2}), indicating that while these methods correctly localize the instances, they fail to assign the correct group. 
Moreover, we find that most models generate only about one instance per sample on average, as illustrated in Figure~\ref{fig: box analysis}.
These observations suggest potential for improvement in generating multiple instances and assigning groups.

Inspired by MS-COCO evaluation~\cite{lin2014microsoft}, we divide the instances into different scales (\ie, small, medium and large) and analyze the performance of different object sizes in Figure~\ref{fig: size analysis}.
We find that while existing models correctly localize large-scale instances, they usually struggle to ground medium and small objects, particularly MLLMs.
The agentic baseline integrates the reasoning capabilities of GPT-4o with the localization strength of G-DNIO, demonstrating significant advantages in grounding small objects.

In order to verify the models' capabilities to reject negative samples, we calculate the average number of predictions across all negative samples, as shown in Figure~\ref{fig: negative analysis}.
We observe that most models struggle with negative samples. 
Gemini~\cite{reid2024gemini} performs the best, with 0.42 predictions per negative sample, but this is still significantly worse than human performance (0.19 predictions per negative sample).
\section{Conclusion}
This paper investigates a valuable yet overlooked problem in the field of MLLMs and proposes a new task, namely multi-context visual grounding.
Unlike prior works that focus on single-image understanding, multi-context visual grounding aims at localizing instances in multi-image scenarios.
Additionally, the text prompts used in multi-context visual grounding are more open-ended and challenging compared to those in previous language-based localization tasks.
To facilitate the research, we introduce MC-Bench, a new benchmark designed for instance-level tasks in multi-context scenarios.
MC-Bench contains 2,000 image pairs with diverse text prompts describing target instances in three distinct styles, covering 20 practical tasks.
After benchmarking over 20 advanced MLLMs and foundation models, we found that current models typically struggle with multiple images and exhibit frustratingly low performance compared to the human upper bound.
We conduct multiple analytical experiments to further investigate the issues that hinder the improvement of existing methods and to identify future directions for development.
Our research advances MLLM development by highlighting weaknesses in instance-level tasks within multi-image scenarios, and MC-Bench serves as a valuable resource for further research. 
We hope our findings will draw attention to the application of MLLMs in instance-level tasks in multi-context scenarios.
\clearpage
\appendix

\section{General Discussions}\label{sec: license and intention}
\subsection{License}\label{sec: license}
MC-Bench dataset is licensed under the Creative Commons Attribution 4.0 International License (CC BY 4.0).
The license applies to all images and annotations we have directly contributed. 
MC-Bench also incorporates images sourced from pre-existing collections. 
For these images, the original licensing terms are respected and remain applicable.

\subsection{Intended Use}
We believe that images in the real world are not isolated, but are inherently linked via spatial, temporal or semantic context.
MC-Bench is initially constructed to facilitate a significant yet largely overlooked research problem, \ie, multi-context visual grounding (grounding objects using open-ended textual prompts in multi-image scenarios).

\begin{figure}[t]
\centering
\includegraphics[width=\linewidth]{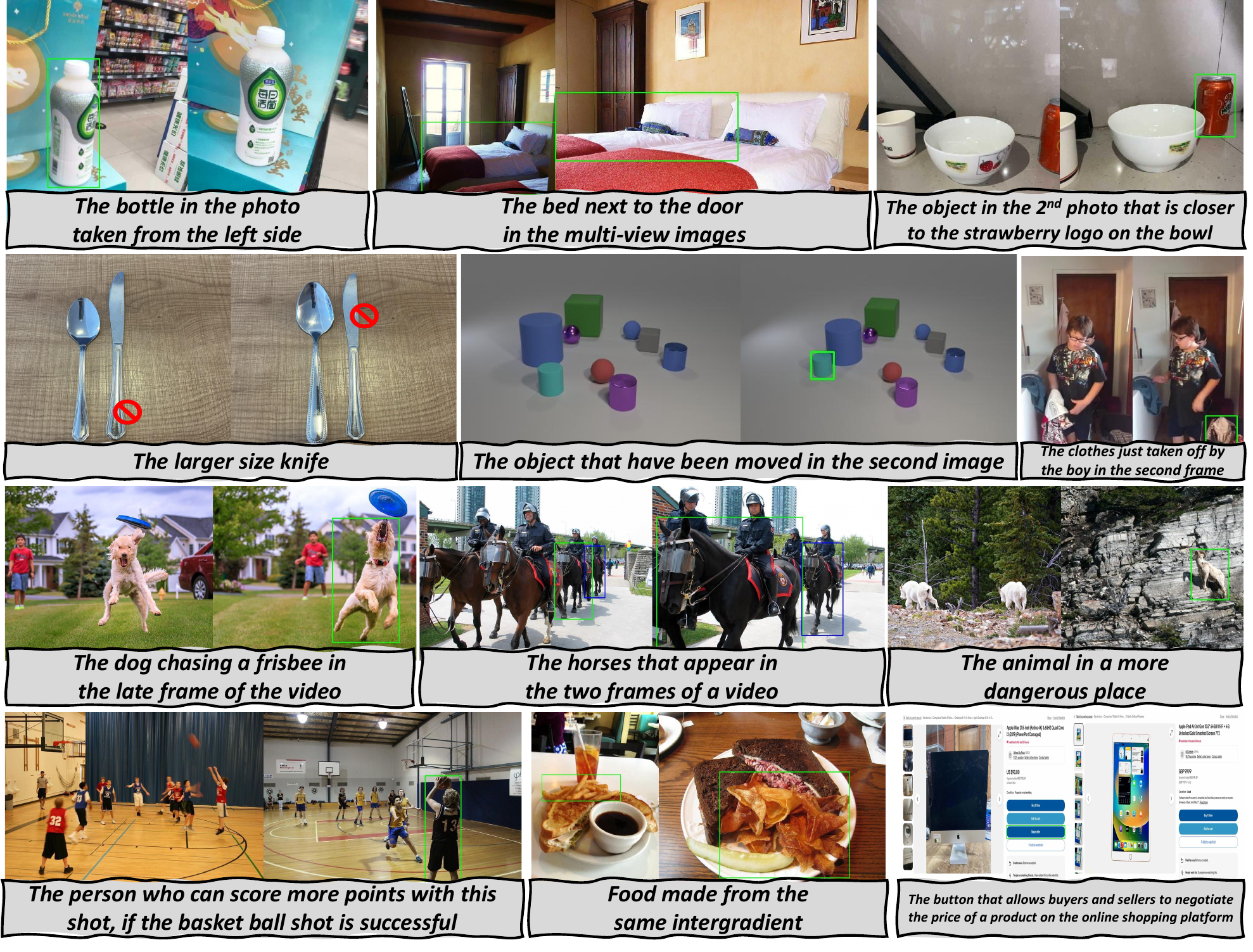}
\vspace{-2.25em}
\caption{MC-Bench includes spatial-, temporal- or semantic-relevant context evaluations, which is highly relevant and beneficial to various downstream real-world applications (\eg, robotics, surveillance systems and general-purpose assistants).}
\label{fig: pratical example} 
\vspace{-1em}
\end{figure}

MC-Bench is highly relevant and beneficial to various downstream applications.
For spatial-relevant context evaluations, MC-Bench assesses multi-view reasoning, which is particularly valuable for robotic navigation and manipulation (illustrated in Figure~\ref{fig: pratical example}).
In addition to these spatial-aware applications, there is a broader range of potential real-world applications that benefit from leveraging temporal and semantic context (\eg, animal/traffic surveillance, sports/food analysis and GUI agents).

The primary purpose of MC-Bench is to function as a dynamic benchmark that continuously evolves and evaluates MLLMs for multi-context visual grounding.
The preliminary results on MC-Bench not only reveal a large performance gap between current MLLMs and humans, but also identify future directions for development through multiple analytical experiments.
We hope MC-Bench can encourage the research community to delve deeper to discover and enhance these untapped potentials of MLLMs in instance-level tasks particularly in multi-image scenarios.

\subsection{Social Impacts}
The data in MC-Bench is not expected to have specific negative impacts.
As the images in MC-Bench are collected from published and publicly available sources, so there are few privacy concerns.
Our text and bounding box annotations do not contain any offensive, insulting or threatening information.
Although a few human annotations could be subjective, we perform cyclic review and multi-round labeling procedures to reduce the bias and ensure the annotation quality.
Beyond the dataset, MC-Bench evaluates a variety of advanced MLLMs and foundation models.
The generated results of these models could be biased or wrong.
The related social impacts on the usage of AI-generated content may apply to our work.
Overall, we consider MC-Bench exhibits minimal negative social impacts.

\subsection{Limitations and Future Works}\label{sec: limitations}
Although MC-Bench evaluates a wide spectrum of potential skills, it does not cover all possible vision-language tasks in real world and exhibits a long-tail distribution. 
Over time, we aim to expand MC-Bench by adding a greater variety of tasks and increasing the number of samples for the tail tasks.
Meanwhile, MC-Bench currently focuses on multi-context samples consisting of two images and one corresponding text description. 
In the future, we aim to extend MC-Bench to accommodate a more general multi-context visual grounding task by incorporating more multi-context samples, each containing a larger number of images.

By the deadline for paper submission, we have evaluated $\sim$20 recent representative approaches with publicly available checkpoints or APIs. 
Since several concurrent works have yet to release their code or checkpoints, we leave their evaluation for future work. 
We plan to establish a leaderboard and update it as new approaches are introduced.

Benchmark results on MC-bench reveal a significant performance gap between MLLMs and humans, especially for the end-to-end models. 
While a few MLLMs accept image sequences as inputs, few of them are specifically designed for instance-level tasks.
Our analysis experiments also show some potential areas for improvement.
Driven by these observations, we plan to investigate more effective solutions for the multi-context visual grounding task in the future.

\section{Implementation Details}\label{sec: imp}

\begin{table*}[t]
\footnotesize
\caption{Existing datasets incorporated in our MC-Bench. We collect and repurpose the images for multi-context visual grounding. The original tasks, original license information and URL links of source datasets are provided.}
\vspace{-1em}
\label{tab: used dataset}
\centering
\setlength{\tabcolsep}{20pt}
\begin{tabular}{lccc}
\toprule
Source Datasets & Original Tasks & Original Licenses & URL Links \\
\midrule
MS-COCO~\cite{lin2014microsoft} & instance segmentation and image captioning  & CC BY 4.0 & \href{https://cocodataset.org/#download}{URL} \\
GRD~\cite{wu2023advancing} & referring expression segmentation & CC BY 4.0 & \href{https://github.com/shikras/d-cube}{URL} \\
Q-Bench~\cite{wu2024qbench} & visual question answering on image quality & CC BY-NC-SA 4.0 & \href{https://huggingface.co/datasets/teowu/LLVisionQA-QBench}{URL} \\
Mantis-Eval~\cite{jiang2024mantis} & multi-image visual question answering & Apache-2.0 & \href{https://huggingface.co/datasets/TIGER-Lab/Mantis-Eval}{URL} \\
DocVQA~\cite{mathew2021docvqa} & visual question answering on documents & N/A & \href{https://www.docvqa.org/datasets/docvqa}{URL} \\
BLINK~\cite{fu2024blink} & question answering on visual perception tasks & Apache-2.0 & \href{https://huggingface.co/datasets/BLINK-Benchmark/BLINK}{URL} \\
CLEVR-Change~\cite{park2019robust} & visual question answering on scene changes & CC BY 4.0 & \href{https://github.com/Seth-Park/RobustChangeCaptioning}{URL} \\
STAR~\cite{wu2024star} & visual question answering on videos & Apache-2.0 & \href{https://bobbywu.com/STAR}{URL} \\
NLVR2~\cite{suhr2019corpus} & multi-image visual question answering & N/A & \href{https://lil.nlp.cornell.edu/nlvr}{URL} \\
WinoGAViL~\cite{NEURIPS2022_a96fe863} & vision-language associations & CC BY 4.0 & \href{https://huggingface.co/datasets/nlphuji/winogavil}{URL} \\
SEED-Bench2-plus~\cite{li2024seed2plus} & visual question answering on text-rich images & CC BY 4.0 & \href{https://huggingface.co/datasets/AILab-CVC/SEED-Bench-2-plus}{URL} \\
\bottomrule
\end{tabular}
\vspace{-1em}
\end{table*}

\subsection{Existing Datasets Incorporated in MC-Bench}\label{sec: existing datasets}
The images in MC-Bench are collected from multiple data sources.
We list the used source datasets in Table~\ref{tab: used dataset}, and we also summarize their original tasks, original license information and URL links that may apply to future users.
For datasets released under licenses other than CC BY (\eg, Q-Bench~\cite{wu2024qbench}), we obtain permission to include their images in our MC-Bench.

\begin{figure}[t]
\centering
\includegraphics[width=\linewidth]{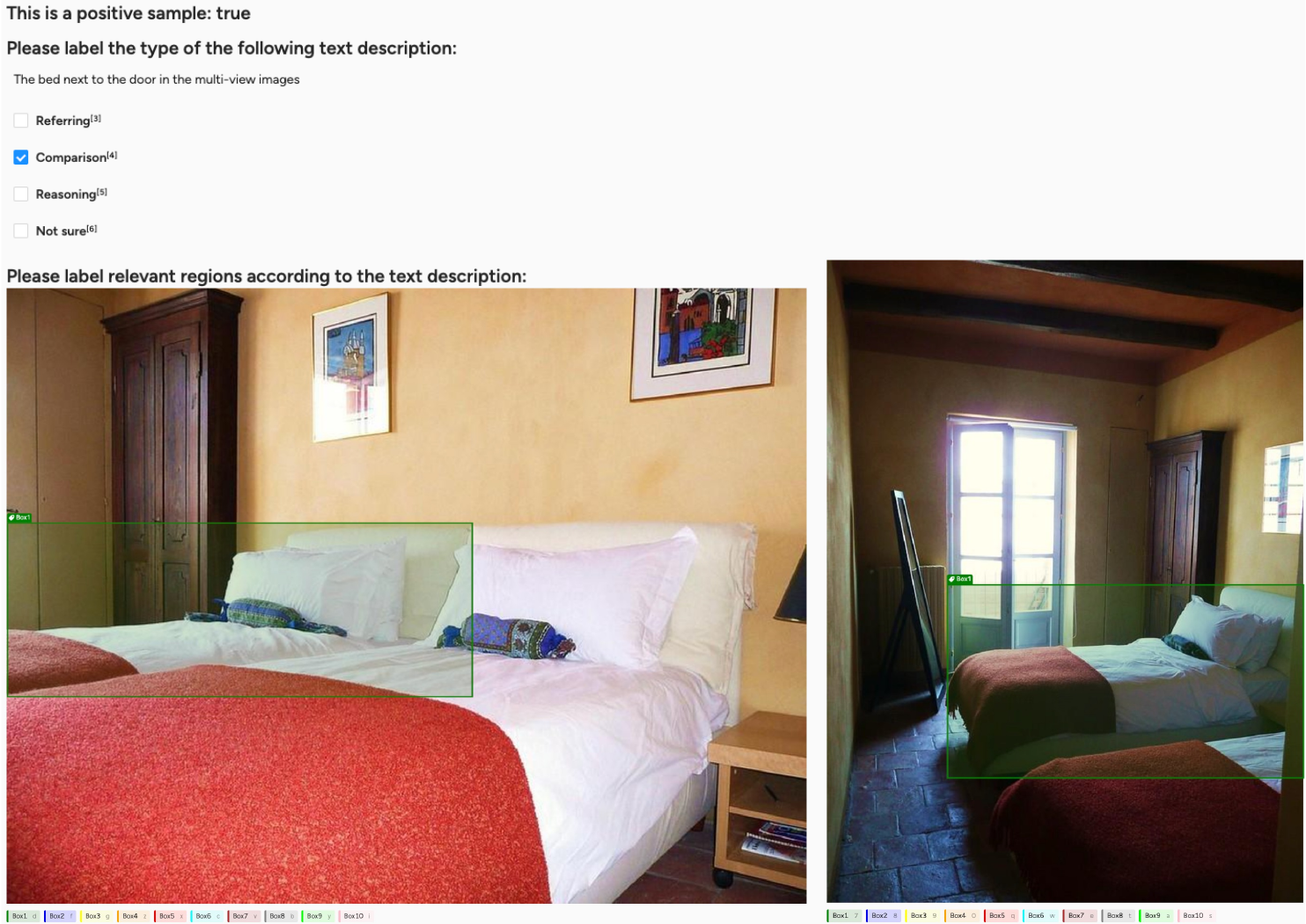}
\vspace{-2em}
\caption{The interface for collecting human annotations.}
\label{fig: annotation} 
\vspace{-1em}
\end{figure}

\subsection{Annotation Interfaces}
We use the open-source annotation tool Label Studio~\cite{LabelStudio} for annotations, in instance-level labeling stage.
Figure~\ref{fig: annotation} shows the user-friendly interface used for collecting human annotations.
The \textit{positive sample label} in top left of the annotation interface indicates whether this sample is a positive sample.
The annotators are first asked to verify whether the positive/negative sample label is correct.
They are then required to categorize the style of the text prompts and draw the bounding boxes.
If \textit{positive sample label} of the sample being annotated is False, no box should be annotated.

\subsection{Annotation Format}
The annotation format of our MC-Bench is similar to MS-COCO~\cite{lin2014microsoft}.
As illustrated in Table~\ref{tab: annotation format}, the main contents are saved in \texttt{description}, \texttt{image} and \texttt{annotation}, including the text prompts, image and instance-level annotation information. 
The annotations are stored using JSON file. 
Our MC-Bench API can be used to access and manipulate annotations.

\begin{table}[t]
\centering
\caption{Annotation format of MC-Bench.}
\vspace{-1.5em}
\label{tab: annotation format}
\begin{tcolorbox}[colback=gray!5!white,colframe=gray!75!black,boxsep=2pt,left=2pt,right=2pt,top=-5pt,bottom=-5pt]
\begin{lstlisting}[basicstyle=\ttfamily\small, frame=none]
{
    "info"           : info,
    "images"         : [image],
    "annotations"    : [annotation],
    "descriptions"   : [description],
    "categories"     : categories
}

description{
    "id"             : int,
    "images_id"      : [int],
    "text"           : str,
    "positive_sample": bool,
    "text_style"     : str
}

image{
    "id"             : int,
    "text_id"        : int,
    "inter_img_id"   : int,
    "file_name"      : str,
    "height"         : int,
    "width"          : int
}

annotation{
    "id"            : int,
    "image_id"      : int,
    "text_id"       : int,
    "category_id"   : int,
    "area"          : int,
    "bbox"          : [x,y,w,h],
    "iscrowd"       : 0 or 1
}
\end{lstlisting}
\end{tcolorbox}
\vspace{-1.25em}
\end{table}

\begin{table*}[t]
\centering
\caption{The prompt we used for GPT-4o.}
\vspace{-1.5em}
\label{tab: gpt prompt}
\begin{tcolorbox}[colback=gray!5!white,colframe=gray!75!black,boxsep=2pt,left=2pt,right=2pt,top=2pt,bottom=2pt]
\textbf{System:}\\
\textit{\# Your Role: excellent object detector\\
\\
\#\# Objective\\
You will be provided with two images and a text describing some instances of interest in the images. Then, you will analyze all inputs and find instances / regions in the images that match the input text prompt from the images. Finally, you will output high-quality bounding box coordinates for each potential instance / region.\\
\\
\#\# Key Guidelines\\
1. Generate one bounding box for one potential instance / region. Do not output bounding boxes covering multiple instances.\\
2. The top-left corner of the input images is coordinate [0, 0], and the bottom-right corner is [1, 1]. The output bounding box coordinate is in [x, y, w, h] format. You should also give confidence scores (range from 0 to 1) for every bounding boxes you predict.\\
3. The input textual prompt can indicate one or more instances / regions within the image pairs, or it can indicate no instance / region.\\
4. You should make full use of contextual information across input images to compare, analyze, and reason to find the target instances.\\
5. Output results strictly in accordance with the given output format, and converted into JSON format.\\
6. The input text prompt may describe multiple groups of instances. For example, ‘apples of the same colors’ may indicate several red apples and several green apples. In such case, you should group the red and green apple into different groups and add addition keys in the output, e.g., ‘Boxes within image1 (group 2)’. In each group, all the apples referred to is either red or green.\\
7. The group number depends on the inputs. The ‘Boxes within ...’ keys are not output if not applicable.\\
\\
\#\# Output Format\\
Input prompt: [textual description for the candidate instances / regions]\\
Analysis: [interpret text prompt and paired images, then explain some key factors for decision making]\\
Positive: [answer ‘True’ if there is any relevant instance, otherwise answer ‘False’]\\
Selected image: [answer ‘image1’ or ‘image2’ or ‘both’ or ‘none’]\\
Boxes within image1 (group 1): [[box1 for instance1], [box2 for instance2], ...]\\
Scores within image1 (group 1) [score1 for box1, score2 for box2, ...]\\
Boxes within image2 (group 1): [[box3 for instance3], [box4 for instance4], ...]\\
Scores within image1 (group 1) [score3 for box3, score4 for box4, ...]\\
Boxes within image1 (group 2): [[box5 for instance5], [box6 for instance6], ...]\\
Scores within image1 (group 2) [score5 for box5, score6 for box6, ...]\\
Boxes within image2 (group 2): [[box7 for instance7], [box8 for instance8], ...]\\
Scores within image1 (group 2) [score7 for box7, score8 for box8, ...] (remove or add more groups if applicable)
}\\
\\
\textbf{User:} $<$input description$>$$<$image 1$>$$<$image 2$>$\\
\textbf{GPT:} \textit{......} \textcolor{Gray}{(bounding box coordinates following given format)}
\end{tcolorbox}
\vspace{-1em}
\end{table*}

\subsection{Evaluated Baselines}
\noindent\textbf{Existing End-to-End Baselines.}
We evaluate several existing models with potential for multi-context visual grounding, including latest proprietary and open-source MLLMs as well as foundation models without LLMs.
All the baselines are evaluated with the official pre-trained models and default hyper-parameters.
For the proprietary API-based models, we evaluate the \texttt{gpt-4o-2024-05-13} version of GPT-4o~\cite{achiam2023gpt} and the \texttt{gemini-1.5-pro-002} version of Gemini~\cite{reid2024gemini}.
All experiments on open-source models were conducted on 4 NVIDIA RTX 3090 GPUs, except for Qwen2-VL-72B~\cite{Qwen2-VL}, which was excluded due to memory constraints.

For the models inherently accept multi-image inputs, we feed the image sequences to the models.
For the models only supports single-image inputs, we horizontally concatenate image pairs and feed the merged images to the models.
To allow the models to distinguish between image pairs, we add a thin white band between two images.

\begin{table*}[t]
\centering
\caption{The multi-round prompt we used for Gemini-1.5-Pro.}
\vspace{-1.5em}
\label{tab: gemini prompt}
\begin{tcolorbox}[colback=gray!5!white,colframe=gray!75!black,boxsep=2pt,left=2pt,right=2pt,top=2pt,bottom=2pt]
\textbf{User:} \textit{Which picture contains the instance described by $<$input description$>$?$<$image 1$>$$<$image 2$>$}\\
\textbf{Gemini:} \textit{......} \textcolor{Gray}{(unstructured results containing image referents and concise analysis)}\\
\\
\textbf{User:} \textit{Please answer using one of \{both pictures, the first picture, the second picture, no existence\}.}\\
\textbf{Gemini:} \textit{......} \textcolor{Gray}{(image referents selected from the given template)}\\
\\
\textbf{User:} \textit{Return bounding boxes for $<$input description$>$, using [ymin, xmin, ymax, xmax] format.$<$image n$>$}\\
\textbf{Gemini:} \textit{......} \textcolor{Gray}{(bounding box coordinates following given format)}
\end{tcolorbox}
\vspace{-1em}
\end{table*}

\begin{table*}[t]
\centering
\caption{The prompt we used for Qwen2-VL.}
\vspace{-1.5em}
\label{tab: qwen2-vl prompt}
\begin{tcolorbox}[colback=gray!5!white,colframe=gray!75!black,boxsep=2pt,left=2pt,right=2pt,top=2pt,bottom=2pt]
\textbf{System:}\\
\textit{\# Your Role: excellent object detector\\
\\
\#\# Objective\\
You will be provided with two images and a text describing some instances of interest in the images. Then, you will analyze all inputs and find instances / regions in the images that match the input text prompt from the images. Finally, you will output high-quality bounding box coordinates for each potential instance / region.\\
\\
\#\# Key Guidelines\\
1. Generate one bounding box for one potential instance / region. Do not output bounding boxes covering multiple instances.\\
2. The input textual prompt can indicate one or more instances / regions within the image pairs, or it can indicate no instance / region.\\
3. You should also give confidence scores (range from 0 to 1) for every bounding boxes you predict.\\
4. You should make full use of contextual information across input images to compare, analyze, and reason to find the target instances.\\
5. Output results strictly in accordance with the given output format.\\
\\
\#\# Output Format\\
The output format should strictly follow the examples:\\
1. $<\mid$img\_id\_start$\mid>$xx$<\mid$img\_id\_end$\mid>$$<\mid$object\_ref\_start$\mid>$xxx$<\mid$object\_ref\_end$\mid>$$<\mid$box\_start$\mid>$(xx,xx),(xx,xx)$<\mid$box\_end\\
$\mid>$$<\mid$score\_start$\mid>$xx$<\mid$score\_end$\mid>$\\
2. $<\mid$img\_id\_start$\mid>$xx$<\mid$img\_id\_end$\mid>$$<\mid$object\_ref\_start$\mid>$xxx$<\mid$object\_ref\_end$\mid>$$<\mid$box\_start$\mid>$(xx,xx),(xx,xx)$<\mid$box\_end\\
$\mid>$$<\mid$score\_start$\mid>$xx$<\mid$score\_end$\mid>$$<\mid$img\_id\_start$\mid>$xx$<\mid$img\_id\_end$\mid>$$<\mid$object\_ref\_start$\mid>$xxx$<\mid$object\_ref\_end$\mid>$$<\mid$box\\
\_start$\mid>$(xx,xx),(xx,xx)$<\mid$box\_end$\mid>$$<\mid$score\_start$\mid>$xx$<\mid$score\_end$\mid>$...\\
3. xxx does not exist.
}\\
\\
\textbf{User:} $<$image 1$>$$<$image 2$>$$<$input description$>$\\
\textbf{Qwen:} \textit{......} \textcolor{Gray}{(bounding box coordinates following given format)}
\end{tcolorbox}
\vspace{-1em}
\end{table*}

For the specialist~\cite{chen2023shikra,peng2023kosmos,wei2023lenna,ma2024groma,li-etal-2024-groundinggpt,you2023ferret,wang2023cogvlm} and a few generalist approaches~\cite{bai2023qwen,lin2023sphinx,chen2023minigpt} with predefined grounding prompts, we utilize their default prompts provided to localize target objects within images. 
As for the generalist models~\cite{achiam2023gpt,reid2024gemini,Qwen2-VL} without predefined grounding prompts, we carefully select the optimal prompts to generate the best results.
Tables~\ref{tab: gpt prompt}-\ref{tab: qwen2-vl prompt} showcase the prompts we use.

\begin{table*}[t]
\centering
\caption{The prompt we used for grounding phrase generation in the agentic baseline.}
\vspace{-1.5em}
\label{tab: agentic prompt}
\begin{tcolorbox}[colback=gray!5!white,colframe=gray!75!black,boxsep=2pt,left=2pt,right=2pt,top=2pt,bottom=2pt]
\textbf{System:}\\
\textit{\# Your Role: excellent referring phrase generator\\
\\
\#\# Objective\\
You will be provided with two images and a text describing some instances of interest in the images. Then, you will analyze all inputs and find instances / regions in the images that match the input text prompt from the images. Finally, you will output high-quality referring phrases for each potential instance / region for subsequent grounding tasks.\\
\\
\#\# Key Guidelines\\
1. Writing a unique referring phrase for each potential instance / region. Do not output a phrase to refer to multiple instances.\\
2. The given referring phrases should be as concise as possible while maintaining sufficient distinctiveness, allowing for easy differentiation of an instance from the image based on the provided referring phrases.\\
3. The input textual prompt can indicate one or more instances / regions within the image pairs, or it can indicate no instance / region.\\
4. The given referring phrase could include the appearance, category and context information of the candidate instances / regions. Any other clues that can better differentiate and identify candidate areas/objects are acceptable.\\
5. The given referring phrase cannot contain cross-image information.\\
6. Output results strictly in accordance with the given output format, and converted into JSON format.\\
7. The input text prompt may describe multiple groups of instances. For example, ‘apples of the same colors’ may indicate several red apples and several green apples. In such case, you should group the red and green apple into different groups and add addition keys in the output, e.g., 'Referring phrases for instances within image1 (group 2)'. In each group, all the apples referred to is either red or green.\\
8. The group number depends on the inputs. The 'Referring phrases ...' keys are not output if not applicable.\\
\\
\#\# Output Format\\
Input prompt: [textual description for the candidate instances / regions]\\
Analysis: [interpret text prompt and paired images, then explain some key factors for decision making]\\
Positive: [answer ‘True’ if there is any relevant instance, otherwise answer ‘False’]\\
Selected image: [answer ‘image1’ or ‘image2’ or ‘both’ or ‘none’]\\
Referring phrases for instances within image1 (group 1): [‘phrase1 for instance1’, ‘phrase2 for instance2’, ...]\\
Referring phrases for instances within image2 (group 1): [‘phrase3 for instance3’, ‘phrase4 for instance4’, ...]\\
Referring phrases for instances within image1 (group 2): [‘phrase5 for instance5’, ‘phrase6 for instance6’, ...]\\
Referring phrases for instances within image2 (group 2): [‘phrase7 for instance7’, ‘phrase8 for instance8’, ...] (remove or add more groups if applicable)
}\\
\\
\textbf{User:} $<$input description$>$$<$image 1$>$$<$image 2$>$\\
\textbf{GPT:} \textit{......} \textcolor{Gray}{(generated phrases for subsequent grounding)}
\end{tcolorbox}
\vspace{-1em}
\end{table*}

\vspace{0.3em}\noindent\textbf{Agentic Baseline.}
Following a divide-and-conquer strategy, we first leverage GPT-4o as a reasoning agent to analyze the target regions and generate some referring phrases that are easier for the detector to understand.
Specifically, we use the GPT API and prompt the model of \texttt{gpt-4o-2024-05-13} version to generate the intermediate results, and the utilized prompt is presented in Table~\ref{tab: agentic prompt}.

We extract the phrase information from the JSON files generated by GPT.
Then, we use the phrases as text query to localize objects from corresponding images.
Concretely, the pre-trained G-DINO~\cite{liu2023grounding} with Swin-B~\cite{liu2021swin} backbone is adopted.
As each GPT-generated phrase only refers one instance within images, we selected the top-1 prediction as the final results. 
Moreover, a confidence threshold of 0.05 is used to filter out the less confident predictions.

\vspace{0.3em}\noindent\textbf{Finetuned Baseline.}
We select the advanced Qwen2-VL-7B~\cite{Qwen2-VL} as our baseline and construct an instruction tuning dataset for performance boosting.
Concretely, the instruction tuning dataset contains two different types of data: multi-context samples for  image-level tasks and instance-level tasks.
We collect $\sim$7.5K multi-context samples from Birds-to-Words~\cite{forbes2019neural} and Multi-VQA~\cite{jiang2024mantis}, as datasets for multi-context image-level tasks (\eg, multi-image captioning and image-level VQA) are already available.
Due to the lack of multi-context instance-level task samples, we synthesize pseudo multi-image samples based on existing detection datasets (\ie, LVIS~\cite{gupta2019lvis} and OmniLabel~\cite{schulter2023omnilabel}).
More specifically, we randomly select two images and generate instructions based on the original object category or referring annotations of the two images, such as `\textit{Output the bounding boxes of $<$category name$>$ in the first image}', `\textit{Output the bounding boxes of $<$object description$>$ in the two images}', and \etc.
We generate $\sim$53K synthetic multi-context instance-level task samples for training.

We finetune Qwen2-VL-7B with the LoRA~\cite{hu2021lora} using LLaMA-Factory~\cite{zheng2024llamafactory} framework.
The model is finetuned using $\sim$60K training samples and bfloat16 format over 3 epochs. 
The learning rate is set to 1e-4 with a cosine annealing scheduler, and the global batch size is set to 32. 
All other settings and hyper-parameters follow the default choices of LLaMA-Factory.
After the model trained, we use the prompt `\textit{Output the bounding boxes of $<$input description$>$}' for multi-context visual grounding.

\section{Additional Experimental Results}~\label{sec: additional result}
\begin{figure}[t]
\centering
\includegraphics[width=\linewidth]{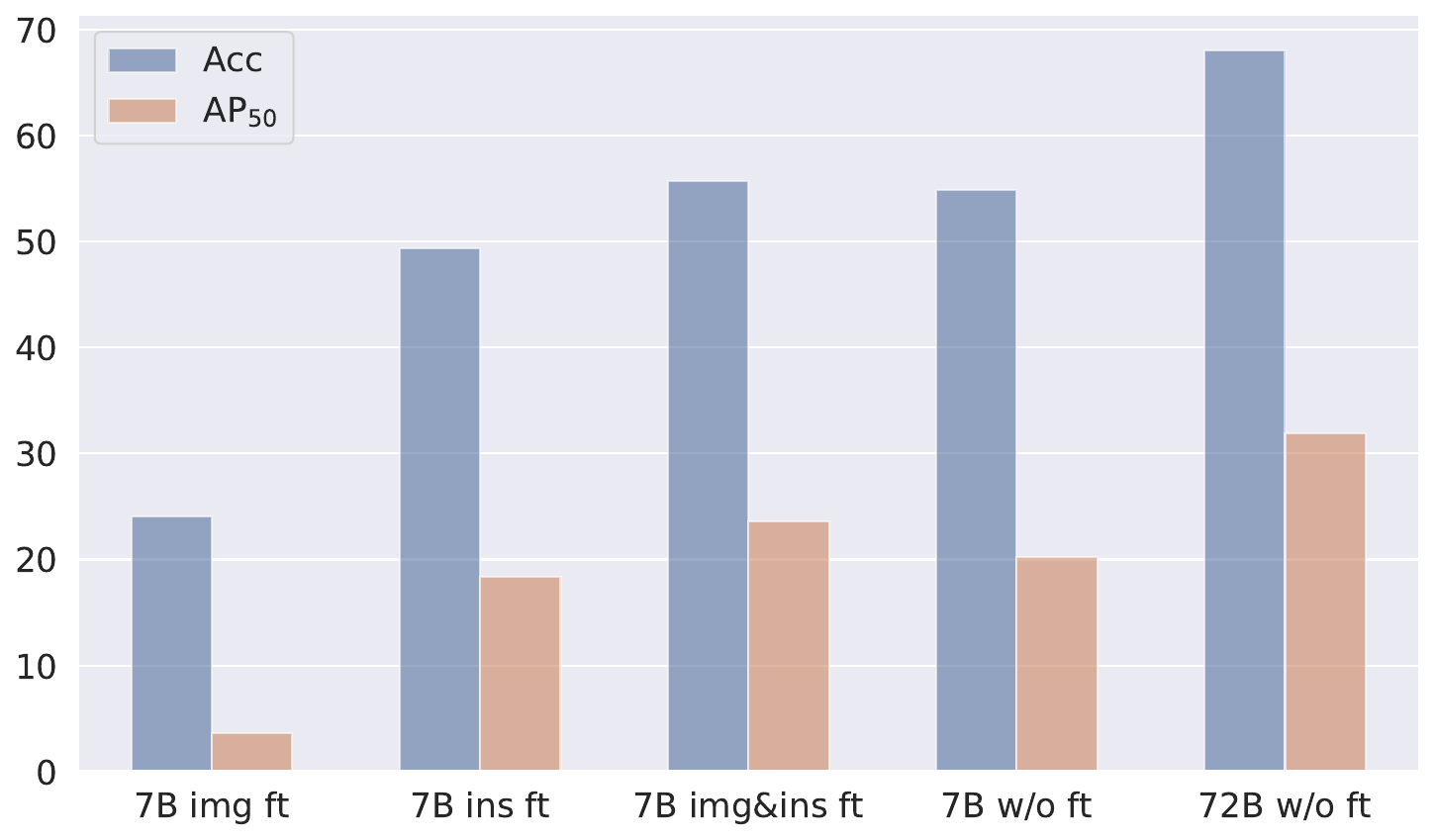}
\vspace{-2.5em}
\caption{Effectiveness of instruction tuning data for the finetuned baseline. The \textit{img ft}, \textit{ins ft} and \textit{img\&ins ft} denote the models trained with collected image-level task samples, synthetic instance-level task samples and merged samples, respectively.}
\label{fig: finetune compare} 
\vspace{-1em}
\end{figure}

\hspace{-0.6em}\noindent\textbf{Ablation Study for the Finetuned Baseline.} 
Figure~\ref{fig: finetune compare} illustrates the effectiveness of instruction tuning with different data. 
Models finetuned with only multi-context image-level task or instance-level task samples obtain performance degradation.
Particularly, the performance of model trained with only collected image-level task samples decreases significantly.
The model trained with only synthetic instance-level task samples also shows sightly performance drop compared to the model without instruction tuning.
We conjecture that most of the synthetic data generated based on object detection datasets~\cite{gupta2019lvis,schulter2023omnilabel} only boosts the cross-image referring abilities and brings limited cross-image comparison and reasoning capabilities. 
After training model with merged data, the finetuned baseline achieves the best performance across image-level and instance-level metrics, surpassing the pre-trained Qwen2-VL-7B by a non-trivial margin.
We also notice that a clear performance gap remains when compared to the 72B model.

\noindent\textbf{Human Evaluation.}
In our current human evaluations, the results of three subjects vary to some extent due to differences in individual cognitive and reasoning levels, as well as the ambiguity and subjectivity of text descriptions. 
We report more detailed human evaluation results in Table~\ref{tab: human compare}. 
We find that the worst performing volunteer still outperforms existing end-to-end MLLMs and achieves on par $\rm{AP_{50}}$ results to the agentic baseline.

\section{Datasheets for MC-Bench}\label{sec: datasheets}
\subsection{Motivation}
\begin{enumerate}
    \item \textbf{For what purpose was the dataset created?} (Was there a specific task in mind? Was there a specific gap that needed to be filled? Please provide a description.)

The primary purpose of MC-Bench is to function as a dynamic benchmark that continuously evolves and evaluates MLLMs for open-ended visual grounding in multi-image scenarios.
This dataset first explores a significant yet largely overlooked research problem, \ie, grounding objects from multi-image inputs based on open-ended textual prompts.
The benchmark results on MC-Bench show a large performance gap between existing MLLMs and humans, as illustrated in 
Table~\ref{tab: performance compare} in the main text.

\item \textbf{Who created this dataset (\eg, which team, research group) and on behalf of which entity (\eg, company, institution, organization)?}

This dataset was created by the authors of this paper.

\item \textbf{Who funded the creation of the dataset?} (If there is an associated grant, please provide the name of the grantor and the grant name and number.)

The institute of the authors funded the creation of the dataset.

\item \textbf{Any other comments?}

None.
\end{enumerate}

\begin{table}[t]
\footnotesize
\caption{Detailed human evaluation results of three volunteers on MC-Bench. }
\label{tab: human compare}
\vspace{-1em}
\centering
\setlength{\tabcolsep}{3pt}
\begin{tabular}{ccccccccc}
\toprule
\# & $\rm{Acc^{ref}}$ & $\rm{Acc^{com}}$ & $\rm{Acc^{rea}}$ & $\rm{Acc}$ & $\rm{AP_{50}^{ref}}$ & $\rm{AP_{50}^{com}}$ & $\rm{AP_{50}^{rea}}$ & $\rm{AP_{50}}$ \\
\midrule
1 & 92.2 & 96.5 & 93.0 & 94.3 & 52.9 & 44.0 & 46.4 & 46.0 \\
2 & 89.6 & 95.6 & 89.9 & 92.2 & 52.6 & 40.7 & 44.1 & 42.8 \\
3 & 86.7 & 94.2 & 88.5 & 90.5 & 37.9 & 36.2 & 32.4 & 35.0 \\
\bottomrule
\end{tabular}
\vspace{-1em}
\end{table}

\subsection{Composition}
\begin{enumerate}

\item \textbf{What do the instances that comprise the dataset represent (\eg, documents, photos, people, countries)?} (Are there multiple types of instances (\eg, movies, users, and ratings; people and interactions between them; nodes and edges)? Please provide a description.)

An instance of our dataset represent the multimodal triplet (\ie, an image pair and a textual prompt describing the regions/objects within the images). 
More detailed descriptions are provided in our paper.

\item \textbf{How many instances are there in total (of each type, if appropriate)?}

Our dataset owns 2,000 samples (\ie, paired images and corresponding text descriptions). 
We provide more detailed dataset statistics in \S\ref{sec: dataset statistics} in the main text.

\item \textbf{Does the dataset contain all possible instances or is it a sample (not necessarily random) of instances from a larger set?} (If the dataset is a sample, then what is the larger set? Is the sample representative of the larger set (\eg, geographic coverage)? If so, please describe how this representativeness was validated/verified. If it is not representative of the larger set, please describe why not (\eg, to cover a more diverse range of instances, because instances were withheld or unavailable).)

The dataset cannot contain all possible instances, as the dataset is designed for open-ended visual grounding evaluation. 
We try to covering diverse range of image domains, disciplines and skills, but we can't guarantee a full sampling of them as discussed in \S\ref{sec: limitations}.

\item \textbf{What data does each instance consist of?} "Raw" data (\eg, unprocessed text or images) or features? In either case, please provide a description

Each instance of our dataset represent an image pair and a textual prompt describing the regions/objects within the images.

\item \textbf{Is there a label or target associated with each instance?} If so, please provide a description.

Yes. We provide the bounding box annotations covering the regions described by the textual prompts. 
More detailed descriptions are provide in our paper.

\item \textbf{Is any information missing from individual instances?} (If so, please provide a description, explaining why this information is missing (\eg, because it was unavailable). This does not include intentionally removed information, but might include, \eg, redacted text.)

No. All necessary information has been provided.

\item \textbf{Are relationships between individual instances made explicit (\eg, users' movie ratings, social network links)?} ( If so, please describe how these relationships are made explicit.)

Yes. Instances are categorized into three groups (\ie, referring, comparison and reasoning) based on the text prompt style of each instance.

\item \textbf{Are there recommended data splits (\eg, training, development/validation, testing)?} (If so, please provide a description of these splits, explaining the rationale behind them.)

Yes. As MC-Bench is an evaluate-only dataset, all samples belong to the testing split.

\item \textbf{Are there any errors, sources of noise, or redundancies in the dataset?} (If so, please provide a description.)

Yes. We try our best to improve the quality of annotations, but the dataset might still contain a few missing labeled objects or subjectivity inconsistencies.

\item \textbf{Is the dataset self-contained, or does it link to or otherwise rely on external resources (\eg, websites, tweets, other datasets)?} (If it links to or relies on external resources, a) are there guarantees that they will exist, and remain constant, over time; b) are there official archival versions of the complete dataset (\ie, including the external resources as they existed at the time the dataset was created); c) are there any restrictions (\eg., licenses, fees) associated with any of the external resources that might apply to a future user? Please provide descriptions of all external resources and any restrictions associated with them, as well as links or other access points, as appropriate.)

Images in MC-Bench are from other publicly available datasets or self-contained. 
We repurpose these images for multi-context visual grounding.
These external datasets are commonly used and long-term exist.
We use the official archival versions of them.
More detailed descriptions of all external resources are provided in \S\ref{sec: existing datasets}.

\item \textbf{Does the dataset contain data that might be considered confidential (\eg, data that is protected by legal privilege or by doctor-patient confidentiality, data that includes the content of individuals' non-public communications)?} (If so, please provide a description.)

No.

\item \textbf{Does the dataset contain data that, if viewed directly, might be offensive, insulting, threatening, or might otherwise cause anxiety?} (If so, please describe why.)

Yes. Some of the scenes may bring anxiety to some people, \eg, photos of car accidents and hospital surgeries.
However, we consider our dataset's offensiveness to be limited, since the source images are collected from prior public datasets.

\item \textbf{Does the dataset identify any subpopulations (\eg, by age, gender)?} (If so, please describe how these subpopulations are identified and provide a description of their respective distributions within the dataset.)

No. This is not explicitly identified. 

\item \textbf{Is it possible to identify individuals (\ie, one or more natural persons), either directly or indirectly (\ie, in combination with other data) from the dataset?} (If so, please describe how.)

Yes. Some samples are about referring expression understanding, where models are required to localize some individuals from images based on the textual descriptions.

\item \textbf{Does the dataset contain data that might be considered sensitive in any way (\eg, data that reveals racial or ethnic origins, sexual orientations, religious beliefs, political opinions or union memberships, or locations; financial or health data; biometric or genetic data; forms of government identification, such as social security numbers; criminal history)?} (If so, please provide a description.)

No. There are no sensitive data used.

\item \textbf{Any other comments?}

None.
\end{enumerate}
\subsection{Collection Process}
\begin{enumerate}
    \item \textbf{How was the data associated with each instance acquired?} (Was the data directly observable (\eg, raw text, movie ratings), reported by subjects (\eg, survey responses), or indirectly inferred/derived from other data (\eg, part-of-speech tags, model-based guesses for age or language)? If data was reported by subjects or indirectly inferred/derived from other data, was the data validated/verified? If so, please describe how.)

    The images are collected from existing public data sources.
    The text descriptions of image pairs are written by the annotators, based on the content of the image pairs.

    \item \textbf{What mechanisms or procedures were used to collect the data (\eg, hardware apparatus or sensor, manual human curation, software program, software API)?} (How were these mechanisms or procedures validated?)

    Software program and manual human curation.
    
    \item \textbf{If the dataset is a sample from a larger set, what was the sampling strategy (\eg, deterministic, probabilistic with specific sampling probabilities)?}
    
    The image are randomly selected from other datasets with specific topics.

    \item \textbf{Who was involved in the data collection process (\eg, students, crowdworkers, contractors) and how were they compensated (\eg, how much were crowdworkers paid)?}

    The data are collected by the authors and students. The involved students are paid nicely.

    \item \textbf{Over what timeframe was the data collected?} (Does this timeframe match the creation timeframe of the data associated with the instances (\eg, recent crawl of old news articles)? If not, please describe the timeframe in which the data associated with the instances was created.)
    
    The dataset was collected in the Spring of 2024, which does not necessarily reflect the timeframe of the data collected.

    \item \textbf{Were any ethical review processes conducted (\eg, by an institutional review board)?} (If so, please provide a description of these review processes, including the outcomes, as well as a link or other access point to any supporting documentation.)

    No ethical review processes were conducted, since the source images are collected from other public datasets.

    \item \textbf{Did you collect the data from the individuals in question directly, or obtain it via third parties or other sources (\eg, websites)?}

    The images are collected from other sources (\ie, repurpose published datasets), while the text descriptions and bounding boxes are labeled by our annotators.

    \item \textbf{Were the individuals in question notified about the data collection?} (If so, please describe (or show with screenshots or other information) how notice was provided, and provide a link or other access point to, or otherwise reproduce, the exact language of the notification itself.)
    
    N/A.

    \item \textbf{Did the individuals in question consent to the collection and use of their data?} (If so, please describe (or show with screenshots or other information) how consent was requested and provided, and provide a link or other access point to, or otherwise reproduce, the exact language to which the individuals consented.)

   N/A.

   \item \textbf{If consent was obtained, were the consenting individuals provided with a mechanism to revoke their consent in the future or for certain uses?} (If so, please provide a description, as well as a link or other access point to the mechanism (if appropriate).)

   N/A.

   \item \textbf{Has an analysis of the potential impact of the dataset and its use on data subjects (\eg, a data protection impact analysis) been conducted?} (If so, please provide a description of this analysis, including the outcomes, as well as a link or other access point to any supporting documentation.)

   N/A. 
   
   \item \textbf{Any other comments?}
  
   None.
\end{enumerate}

\subsection{Preprocessing/cleaning/labeling}
\begin{enumerate}
    \item \textbf{Was any preprocessing/cleaning/labeling of the data done (\eg, discretization or bucketing, tokenization, part-of-speech tagging, SIFT feature extraction, removal of instances, processing of missing values)?} (If so, please provide a description. If not, you may skip the remainder of the questions in this section.)

    Yes. We reorganized images collected from existing datasets and introduced extra annotations.
    Specifically, we provided textual prompts for each image pair describing some objects within the images, and we also labeled the language-grounded regions using bounding boxes.
    
    \item \textbf{Was the “raw” data saved in addition to the preprocessed/cleaned/labeled data (\eg, to support unanticipated future uses)?} If so, please provide a link or other access point to the “raw” data.

    Yes. MC-Bench itself contains partial the raw data (\ie, textual descriptions and bounding box annotations). 
    The rest of raw data (\ie, images) were collected from other published datasets (see \S\ref{sec: existing datasets}) and we did not modify the images.

    \item \textbf{Is the software that was used to preprocess/clean/label the data available?} If so, please provide a link or other access point.
    
    We leverage the open-source annotation tool, Label Studio (\url{https://github.com/HumanSignal/label-studio}), in both text and box annotation stages, owing to its programmable and user-friendly interface for annotating paired images.

    \item \textbf{Any other comments?}
    
    None.

\end{enumerate}
\subsection{Uses}

\begin{enumerate}
    \item \textbf{Has the dataset been used for any tasks already?} (If so, please provide a description.)

	The images of MC-Bench are collected from published datasets for other tasks. 
	In contrast, the textual prompts and bounding box annotations in MC-Bench are newly introduced and have not used for any other tasks.

    \item \textbf{Is there a repository that links to any or all papers or systems that use the dataset?} (If so, please provide a link or other access point.)

    Yes. We are going to maintain a leaderboard for MC-Bench on the project page (\url{https://xuyunqiu.github.io/MC-Bench}). 
    The links of all the evaluated methods will be provided.

    \item \textbf{What (other) tasks could the dataset be used for?}

    There are many more, such as multi-image VQA and common object detection.

    \item \textbf{Is there anything about the composition of the dataset or the way it was collected and preprocessed/cleaned/labeled that might impact future uses?} (For example, is there anything that a future user might need to know to avoid uses that could result in unfair treatment of individuals or groups (\eg, stereotyping, quality of service issues) or other undesirable harms (\eg, financial harms, legal risks) If so, please provide a description. Is there anything a future user could do to mitigate these undesirable harms?)

    No.

    \item \textbf{Are there tasks for which the dataset should not be used?} (If so, please provide a description.)

    No.

    \item \textbf{Any other comments?}

    None.
\end{enumerate}

\subsection{Distribution}

\begin{enumerate}
    \item \textbf{Will the dataset be distributed to third parties outside of the entity (\eg, company, institution, organization) on behalf of which the dataset was created?} (If so, please provide a description.)

    Yes, the dataset is publicly available on the Internet.

    \item \textbf{How will the dataset will be distributed (\eg, tarball on website, API, GitHub)?} (Does the dataset have a digital object identifier (DOI)?)

    On our GitHub project page (\url{https://xuyunqiu.github.io/MC-Bench}).

    \item \textbf{When will the dataset be distributed?}

    The dataset was first released in June 2024.
    The updated dataset (MC-Bench v0.5) will be released alongside the ICCV camera-ready version in late July 2025.

    \item \textbf{Will the dataset be distributed under a copyright or other intellectual property (IP) license, and/or under applicable terms of use (ToU)?} (If so, please describe this license and/or ToU, and provide a link or other access point to, or otherwise reproduce, any relevant licensing terms or ToU, as well as any fees associated with these restrictions.)

    The dataset is licensed under a CC license. More detailed license information is provided in \S\ref{sec: license}.

    \item \textbf{Have any third parties imposed IP-based or other restrictions on the data associated with the instances?} (If so, please describe these restrictions, and provide a link or other access point to, or otherwise reproduce, any relevant licensing terms, as well as any fees associated with these restrictions.)

    As far as we know, no.

    \item \textbf{Do any export controls or other regulatory restrictions apply to the dataset or to individual instances?} (If so, please describe these restrictions, and provide a link or other access point to, or otherwise reproduce, any supporting documentation.)

    As far as we know, no.

    \item \textbf{Any other comments?}

    None.
\end{enumerate}
\subsection{Maintenance}

\begin{enumerate}
    \item \textbf{Who is supporting/hosting/maintaining the dataset?}

    The authors.

    \item \textbf{How can the owner/curator/manager of the dataset be contacted (\eg, email address)?}

    The dataset owner can be contacted through the authors' email address.

    \item \textbf{Is there an erratum?} (If so, please provide a link or other access point.)

    Currently, no. As errors are encountered, future versions of the dataset may be released (but will be versioned). 

    \item \textbf{Will the dataset be updated (\eg, to correct labeling errors, add new instances, delete instances')?} (If so, please describe how often, by whom, and how updates will be communicated to users (\eg, mailing list, GitHub)?)

    Yes. The dataset will be updated by the dataset owner. The update information will be posted on the project page.
    
    \item \textbf{If the dataset relates to people, are there applicable limits on the retention of the data associated with the instances (\eg, were individuals in question told that their data would be retained for a fixed period of time and then deleted)?} (If so, please describe these limits and explain how they will be enforced.)

    No.
    
    \item \textbf{Will older versions of the dataset continue to be supported/hosted/maintained?} (If so, please describe how. If not, please describe how its obsolescence will be communicated to users.)

    Yes. The older versions of the dataset will be provided in the same webpage.

    \item \textbf{If others want to extend/augment/build on/contribute to the dataset, is there a mechanism for them to do so?} (If so, please provide a description. Will these contributions be validated/verified? If so, please describe how. If not, why not? Is there a process for communicating/distributing these contributions to other users? If so, please provide a description.)

    Yes. Others may do so and should contact the original authors about incorporating fixes/extensions.

    \item \textbf{Any other comments?}

    None.
\end{enumerate}

{
    \small
    \bibliographystyle{ieeenat_fullname}
    \bibliography{main}

\begin{thebibliography}{119}
\providecommand{\natexlab}[1]{#1}
\providecommand{\url}[1]{\texttt{#1}}
\expandafter\ifx\csname urlstyle\endcsname\relax
  \providecommand{\doi}[1]{doi: #1}\else
  \providecommand{\doi}{doi: \begingroup \urlstyle{rm}\Url}\fi

\bibitem[Achiam et~al.(2023)Achiam, Adler, Agarwal, Ahmad, Akkaya, Aleman, Almeida, Altenschmidt, Altman, Anadkat, et~al.]{achiam2023gpt}
Josh Achiam, Steven Adler, Sandhini Agarwal, Lama Ahmad, Ilge Akkaya, Florencia~Leoni Aleman, Diogo Almeida, Janko Altenschmidt, Sam Altman, Shyamal Anadkat, et~al.
\newblock Gpt-4 technical report.
\newblock \emph{arXiv preprint arXiv:2303.08774}, 2023.

\bibitem[Alayrac et~al.(2022)Alayrac, Donahue, Luc, Miech, Barr, Hasson, Lenc, Mensch, Millican, Reynolds, et~al.]{alayrac2022flamingo}
Jean-Baptiste Alayrac, Jeff Donahue, Pauline Luc, Antoine Miech, Iain Barr, Yana Hasson, Karel Lenc, Arthur Mensch, Katherine Millican, Malcolm Reynolds, et~al.
\newblock Flamingo: a visual language model for few-shot learning.
\newblock In \emph{NeurIPS}, 2022.

\bibitem[Antol et~al.(2015)Antol, Agrawal, Lu, Mitchell, Batra, Zitnick, and Parikh]{antol2015vqa}
Stanislaw Antol, Aishwarya Agrawal, Jiasen Lu, Margaret Mitchell, Dhruv Batra, C~Lawrence Zitnick, and Devi Parikh.
\newblock Vqa: Visual question answering.
\newblock In \emph{ICCV}, 2015.

\bibitem[Awadalla et~al.(2023)Awadalla, Gao, Gardner, Hessel, Hanafy, Zhu, Marathe, Bitton, Gadre, Sagawa, et~al.]{awadalla2023openflamingo}
Anas Awadalla, Irena Gao, Josh Gardner, Jack Hessel, Yusuf Hanafy, Wanrong Zhu, Kalyani Marathe, Yonatan Bitton, Samir Gadre, Shiori Sagawa, et~al.
\newblock Openflamingo: An open-source framework for training large autoregressive vision-language models.
\newblock \emph{arXiv preprint arXiv:2308.01390}, 2023.

\bibitem[Bai et~al.(2023)Bai, Bai, Yang, Wang, Tan, Wang, Lin, Zhou, and Zhou]{bai2023qwen}
Jinze Bai, Shuai Bai, Shusheng Yang, Shijie Wang, Sinan Tan, Peng Wang, Junyang Lin, Chang Zhou, and Jingren Zhou.
\newblock Qwen-vl: A frontier large vision-language model with versatile abilities.
\newblock \emph{arXiv preprint arXiv:2308.12966}, 2023.

\bibitem[Bitton et~al.(2022)Bitton, Bitton~Guetta, Yosef, Elovici, Bansal, Stanovsky, and Schwartz]{NEURIPS2022_a96fe863}
Yonatan Bitton, Nitzan Bitton~Guetta, Ron Yosef, Yuval Elovici, Mohit Bansal, Gabriel Stanovsky, and Roy Schwartz.
\newblock Winogavil: Gamified association benchmark to challenge vision-and-language models.
\newblock In \emph{NeurIPS}, 2022.

\bibitem[Cai et~al.(2025)Cai, Ke, Jahangard, de~la Banda, Haffari, Stuckey, and Rezatofighi]{cai2025naver}
Zhixi Cai, Fucai Ke, Simindokht Jahangard, Maria~Garcia de~la Banda, Reza Haffari, Peter~J Stuckey, and Hamid Rezatofighi.
\newblock Naver: A neuro-symbolic compositional automaton for visual grounding with explicit logic reasoning.
\newblock In \emph{ICCV}, 2025.

\bibitem[Carion et~al.(2020)Carion, Massa, Synnaeve, Usunier, Kirillov, and Zagoruyko]{carion2020end}
Nicolas Carion, Francisco Massa, Gabriel Synnaeve, Nicolas Usunier, Alexander Kirillov, and Sergey Zagoruyko.
\newblock End-to-end object detection with transformers.
\newblock In \emph{ECCV}, 2020.

\bibitem[Chen et~al.(2022)Chen, Anjum, and Gurari]{chen2022grounding}
Chongyan Chen, Samreen Anjum, and Danna Gurari.
\newblock Grounding answers for visual questions asked by visually impaired people.
\newblock In \emph{CVPR}, 2022.

\bibitem[Chen et~al.(2025{\natexlab{a}})Chen, Yuan, Xu, Feng, Cen, Liu, Huang, and Yang]{chen2025mathflow}
Felix Chen, Hangjie Yuan, Yunqiu Xu, Tao Feng, Jun Cen, Pengwei Liu, Zeying Huang, and Yi Yang.
\newblock Mathflow: Enhancing the perceptual flow of mllms for visual mathematical problems.
\newblock \emph{arXiv preprint arXiv:2503.16549}, 2025{\natexlab{a}}.

\bibitem[Chen et~al.(2023{\natexlab{a}})Chen, Zhu, Shen, Li, Liu, Zhang, Krishnamoorthi, Chandra, Xiong, and Elhoseiny]{chen2023minigpt}
Jun Chen, Deyao Zhu, Xiaoqian Shen, Xiang Li, Zechun Liu, Pengchuan Zhang, Raghuraman Krishnamoorthi, Vikas Chandra, Yunyang Xiong, and Mohamed Elhoseiny.
\newblock Minigpt-v2: Large language model as a unified interface for vision-language multi-task learning.
\newblock \emph{arXiv preprint arXiv:2310.09478}, 2023{\natexlab{a}}.

\bibitem[Chen et~al.(2024{\natexlab{a}})Chen, Lv, Wu, Lin, Song, Gao, Liu, Gao, Mao, and Shou]{chen2024videollm}
Joya Chen, Zhaoyang Lv, Shiwei Wu, Kevin~Qinghong Lin, Chenan Song, Difei Gao, Jia-Wei Liu, Ziteng Gao, Dongxing Mao, and Mike~Zheng Shou.
\newblock Videollm-online: Online video large language model for streaming video.
\newblock In \emph{CVPR}, 2024{\natexlab{a}}.

\bibitem[Chen et~al.(2025{\natexlab{b}})Chen, Wei, Zhao, Song, Wu, Peng, Chan, and Zhang]{chen2024revisiting}
Jierun Chen, Fangyun Wei, Jinjing Zhao, Sizhe Song, Bohuai Wu, Zhuoxuan Peng, S.-H.~Gary Chan, and Hongyang Zhang.
\newblock Revisiting referring expression comprehension evaluation in the era of large multimodal models.
\newblock In \emph{CVPRW}, 2025{\natexlab{b}}.

\bibitem[Chen et~al.(2023{\natexlab{b}})Chen, Zhang, Zeng, Zhang, Zhu, and Zhao]{chen2023shikra}
Keqin Chen, Zhao Zhang, Weili Zeng, Richong Zhang, Feng Zhu, and Rui Zhao.
\newblock Shikra: Unleashing multimodal llm's referential dialogue magic.
\newblock \emph{arXiv preprint arXiv:2306.15195}, 2023{\natexlab{b}}.

\bibitem[Chen et~al.(2024{\natexlab{b}})Chen, Wang, Cao, Liu, Gao, Cui, Zhu, Ye, Tian, Liu, et~al.]{chen2024expanding}
Zhe Chen, Weiyun Wang, Yue Cao, Yangzhou Liu, Zhangwei Gao, Erfei Cui, Jinguo Zhu, Shenglong Ye, Hao Tian, Zhaoyang Liu, et~al.
\newblock Expanding performance boundaries of open-source multimodal models with model, data, and test-time scaling.
\newblock \emph{arXiv preprint arXiv:2412.05271}, 2024{\natexlab{b}}.

\bibitem[Dai et~al.(2023)Dai, Li, Li, Tiong, Zhao, Wang, Li, Fung, and Hoi]{NEURIPS2023_9a6a435e}
Wenliang Dai, Junnan Li, Dongxu Li, Anthony Tiong, Junqi Zhao, Weisheng Wang, Boyang Li, Pascale~N Fung, and Steven Hoi.
\newblock Instructblip: Towards general-purpose vision-language models with instruction tuning.
\newblock In \emph{NeurIPS}, 2023.

\bibitem[Dhamija et~al.(2020)Dhamija, Gunther, Ventura, and Boult]{dhamija2020overlooked}
Akshay Dhamija, Manuel Gunther, Jonathan Ventura, and Terrance Boult.
\newblock The overlooked elephant of object detection: Open set.
\newblock In \emph{WACV}, 2020.

\bibitem[Ding et~al.(2024)Ding, Han, Xu, Liang, Zhang, and Li]{ding2024holistic}
Xinpeng Ding, Jianhua Han, Hang Xu, Xiaodan Liang, Wei Zhang, and Xiaomeng Li.
\newblock Holistic autonomous driving understanding by bird's-eye-view injected multi-modal large models.
\newblock In \emph{CVPR}, 2024.

\bibitem[Forbes et~al.(2019)Forbes, Kaeser-Chen, Sharma, and Belongie]{forbes2019neural}
Maxwell Forbes, Christine Kaeser-Chen, Piyush Sharma, and Serge Belongie.
\newblock Neural naturalist: Generating fine-grained image comparisons.
\newblock In \emph{EMNLP}, 2019.

\bibitem[Fu et~al.(2024)Fu, Hu, Li, Feng, Wang, Lin, Roth, Smith, Ma, and Krishna]{fu2024blink}
Xingyu Fu, Yushi Hu, Bangzheng Li, Yu Feng, Haoyu Wang, Xudong Lin, Dan Roth, Noah~A Smith, Wei-Chiu Ma, and Ranjay Krishna.
\newblock Blink: Multimodal large language models can see but not perceive.
\newblock In \emph{ECCV}, 2024.

\bibitem[Gan et~al.(2017)Gan, Li, Li, Sun, and Gong]{gan2017vqs}
Chuang Gan, Yandong Li, Haoxiang Li, Chen Sun, and Boqing Gong.
\newblock Vqs: Linking segmentations to questions and answers for supervised attention in vqa and question-focused semantic segmentation.
\newblock In \emph{ICCV}, 2017.

\bibitem[Gao et~al.(2023)Gao, Han, Zhang, Lin, Geng, Zhou, Zhang, Lu, He, Yue, et~al.]{gao2023llama}
Peng Gao, Jiaming Han, Renrui Zhang, Ziyi Lin, Shijie Geng, Aojun Zhou, Wei Zhang, Pan Lu, Conghui He, Xiangyu Yue, et~al.
\newblock Llama-adapter v2: Parameter-efficient visual instruction model.
\newblock \emph{arXiv preprint arXiv:2304.15010}, 2023.

\bibitem[Gu et~al.(2022)Gu, Lin, Kuo, and Cui]{gu2021open}
Xiuye Gu, Tsung-Yi Lin, Weicheng Kuo, and Yin Cui.
\newblock Open-vocabulary object detection via vision and language knowledge distillation.
\newblock In \emph{ICLR}, 2022.

\bibitem[Guo et~al.(2024)Guo, De~Mello, Yin, Byeon, Cheung, Yu, Luo, and Liu]{guo2024regiongpt}
Qiushan Guo, Shalini De~Mello, Hongxu Yin, Wonmin Byeon, Ka~Chun Cheung, Yizhou Yu, Ping Luo, and Sifei Liu.
\newblock Regiongpt: Towards region understanding vision language model.
\newblock In \emph{CVPR}, 2024.

\bibitem[Gupta et~al.(2019)Gupta, Dollar, and Girshick]{gupta2019lvis}
Agrim Gupta, Piotr Dollar, and Ross Girshick.
\newblock Lvis: A dataset for large vocabulary instance segmentation.
\newblock In \emph{CVPR}, 2019.

\bibitem[Gurari et~al.(2018)Gurari, Li, Stangl, Guo, Lin, Grauman, Luo, and Bigham]{gurari2018vizwiz}
Danna Gurari, Qing Li, Abigale~J Stangl, Anhong Guo, Chi Lin, Kristen Grauman, Jiebo Luo, and Jeffrey~P Bigham.
\newblock Vizwiz grand challenge: Answering visual questions from blind people.
\newblock In \emph{CVPR}, 2018.

\bibitem[Han et~al.(2024)Han, Zhu, Lao, and Jiang]{han2023zero}
Zeyu Han, Fangrui Zhu, Qianru Lao, and Huaizu Jiang.
\newblock Zero-shot referring expression comprehension via structural similarity between images and captions.
\newblock In \emph{CVPR}, 2024.

\bibitem[Hu et~al.(2022)Hu, Shen, Wallis, Allen-Zhu, Li, Wang, Wang, and Chen]{hu2021lora}
Edward~J Hu, Yelong Shen, Phillip Wallis, Zeyuan Allen-Zhu, Yuanzhi Li, Shean Wang, Lu Wang, and Weizhu Chen.
\newblock {LoRA}: Low-rank adaptation of large language models.
\newblock In \emph{ICLR}, 2022.

\bibitem[Hudson and Manning(2019)]{hudson2019gqa}
Drew~A Hudson and Christopher~D Manning.
\newblock Gqa: A new dataset for real-world visual reasoning and compositional question answering.
\newblock In \emph{CVPR}, 2019.

\bibitem[Jia et~al.(2024)Jia, Xu, Zhu, Chen, Wang, and Yang]{jia2024mos2}
Heng Jia, Yunqiu Xu, Linchao Zhu, Guang Chen, Yufei Wang, and Yi Yang.
\newblock Mos2: Mixture of scale and shift experts for text-only video captioning.
\newblock In \emph{ACM MM}, 2024.

\bibitem[Jiang et~al.(2024)Jiang, He, Zeng, Wei, Ku, Liu, and Chen]{jiang2024mantis}
Dongfu Jiang, Xuan He, Huaye Zeng, Cong Wei, Max Ku, Qian Liu, and Wenhu Chen.
\newblock Mantis: Interleaved multi-image instruction tuning.
\newblock \emph{TMLR}, 2024.

\bibitem[Jiang et~al.(2019)Jiang, Liang, Chen, Zhu, and Li]{jiang2019class}
Shuqiang Jiang, Sisi Liang, Chengpeng Chen, Yaohui Zhu, and Xiangyang Li.
\newblock Class agnostic image common object detection.
\newblock \emph{IEEE TIP}, 2019.

\bibitem[Jiao et~al.(2024)Jiao, Chen, Jie, Chen, Ma, and Jiang]{jiao2025lumen}
Yang Jiao, Shaoxiang Chen, Zequn Jie, Jingjing Chen, Lin Ma, and Yu-Gang Jiang.
\newblock Lumen: Unleashing versatile vision-centric capabilities of large multimodal models.
\newblock In \emph{NeurIPS}, 2024.

\bibitem[Joseph et~al.(2021)Joseph, Khan, Khan, and Balasubramanian]{joseph2021towards}
KJ Joseph, Salman Khan, Fahad~Shahbaz Khan, and Vineeth~N Balasubramanian.
\newblock Towards open world object detection.
\newblock In \emph{CVPR}, 2021.

\bibitem[Kaplan et~al.(2020)Kaplan, McCandlish, Henighan, Brown, Chess, Child, Gray, Radford, Wu, and Amodei]{kaplan2020scaling}
Jared Kaplan, Sam McCandlish, Tom Henighan, Tom~B Brown, Benjamin Chess, Rewon Child, Scott Gray, Alec Radford, Jeffrey Wu, and Dario Amodei.
\newblock Scaling laws for neural language models.
\newblock \emph{arXiv preprint arXiv:2001.08361}, 2020.

\bibitem[Kazemzadeh et~al.(2014)Kazemzadeh, Ordonez, Matten, and Berg]{kazemzadeh2014referitgame}
Sahar Kazemzadeh, Vicente Ordonez, Mark Matten, and Tamara Berg.
\newblock Referitgame: Referring to objects in photographs of natural scenes.
\newblock In \emph{EMNLP}, 2014.

\bibitem[Kirillov et~al.(2023)Kirillov, Mintun, Ravi, Mao, Rolland, Gustafson, Xiao, Whitehead, Berg, Lo, et~al.]{kirillov2023segment}
Alexander Kirillov, Eric Mintun, Nikhila Ravi, Hanzi Mao, Chloe Rolland, Laura Gustafson, Tete Xiao, Spencer Whitehead, Alexander~C Berg, Wan-Yen Lo, et~al.
\newblock Segment anything.
\newblock In \emph{CVPR}, 2023.

\bibitem[Lauren{\c{c}}on et~al.(2024)Lauren{\c{c}}on, Tronchon, Cord, and Sanh]{laurenccon2025matters}
Hugo Lauren{\c{c}}on, L{\'e}o Tronchon, Matthieu Cord, and Victor Sanh.
\newblock What matters when building vision-language models?
\newblock In \emph{NeurIPS}, 2024.

\bibitem[Li et~al.(2024{\natexlab{a}})Li, Ge, Chen, Ge, Zhang, and Shan]{li2024seed2plus}
Bohao Li, Yuying Ge, Yi Chen, Yixiao Ge, Ruimao Zhang, and Ying Shan.
\newblock Seed-bench-2-plus: Benchmarking multimodal large language models with text-rich visual comprehension.
\newblock \emph{arXiv preprint arXiv:2404.16790}, 2024{\natexlab{a}}.

\bibitem[Li et~al.(2024{\natexlab{b}})Li, Ge, Ge, Wang, Wang, Zhang, and Shan]{li2024seed}
Bohao Li, Yuying Ge, Yixiao Ge, Guangzhi Wang, Rui Wang, Ruimao Zhang, and Ying Shan.
\newblock Seed-bench: Benchmarking multimodal large language models.
\newblock In \emph{CVPR}, 2024{\natexlab{b}}.

\bibitem[Li et~al.(2022)Li, Liu, Li, Zhang, Aneja, Yang, Jin, Hu, Liu, Lee, and Gao]{NEURIPS2022_3c4688b6}
Chunyuan Li, Haotian Liu, Liunian Li, Pengchuan Zhang, Jyoti Aneja, Jianwei Yang, Ping Jin, Houdong Hu, Zicheng Liu, Yong~Jae Lee, and Jianfeng Gao.
\newblock Elevater: A benchmark and toolkit for evaluating language-augmented visual models.
\newblock In \emph{NeurIPS}, 2022.

\bibitem[Li et~al.(2025{\natexlab{a}})Li, Zhang, Zhang, Zhang, Li, Li, Ma, and Li]{li2024llava}
Feng Li, Renrui Zhang, Hao Zhang, Yuanhan Zhang, Bo Li, Wei Li, Zejun Ma, and Chunyuan Li.
\newblock Llava-interleave: Tackling multi-image, video, and 3d in large multimodal models.
\newblock In \emph{ICLR}, 2025{\natexlab{a}}.

\bibitem[Li et~al.(2023{\natexlab{a}})Li, Li, Savarese, and Hoi]{li2023blip}
Junnan Li, Dongxu Li, Silvio Savarese, and Steven Hoi.
\newblock Blip-2: Bootstrapping language-image pre-training with frozen image encoders and large language models.
\newblock In \emph{ICML}, 2023{\natexlab{a}}.

\bibitem[Li et~al.(2024{\natexlab{c}})Li, Pan, Ge, Gao, Zhang, Ji, Zhang, Chua, Tang, and Zhuang]{li2023empowering}
Juncheng Li, Kaihang Pan, Zhiqi Ge, Minghe Gao, Hanwang Zhang, Wei Ji, Wenqiao Zhang, Tat-Seng Chua, Siliang Tang, and Yueting Zhuang.
\newblock Empowering vision-language models to follow interleaved vision-language instructions.
\newblock In \emph{ICLR}, 2024{\natexlab{c}}.

\bibitem[Li et~al.(2021)Li, Guo, and Wang]{li2021proposal}
Kun Li, Dan Guo, and Meng Wang.
\newblock Proposal-free video grounding with contextual pyramid network.
\newblock In \emph{AAAI}, 2021.

\bibitem[Li et~al.(2023{\natexlab{b}})Li, Li, Guo, Yang, and Wang]{li2023transformer}
Kun Li, Jiaxiu Li, Dan Guo, Xun Yang, and Meng Wang.
\newblock Transformer-based visual grounding with cross-modality interaction.
\newblock \emph{TOMM}, 2023{\natexlab{b}}.

\bibitem[Li et~al.(2024{\natexlab{d}})Li, Wang, He, Li, Wang, Liu, Wang, Xu, Chen, Luo, et~al.]{li2023mvbench}
Kunchang Li, Yali Wang, Yinan He, Yizhuo Li, Yi Wang, Yi Liu, Zun Wang, Jilan Xu, Guo Chen, Ping Luo, et~al.
\newblock Mvbench: A comprehensive multi-modal video understanding benchmark.
\newblock In \emph{CVPR}, 2024{\natexlab{d}}.

\bibitem[Li et~al.(2025{\natexlab{b}})Li, Huang, Chen, Huang, Huang, Guo, Liu, Xu, Li, Li, et~al.]{li2025migician}
You Li, Heyu Huang, Chi Chen, Kaiyu Huang, Chao Huang, Zonghao Guo, Zhiyuan Liu, Jinan Xu, Yuhua Li, Ruixuan Li, et~al.
\newblock Migician: Revealing the magic of free-form multi-image grounding in multimodal large language models.
\newblock \emph{arXiv preprint arXiv:2501.05767}, 2025{\natexlab{b}}.

\bibitem[Li et~al.(2024{\natexlab{e}})Li, Xu, Zhang, Song, Cai, Qi, Zhou, Pan, Li, Tu, Huang, and Wang]{li-etal-2024-groundinggpt}
Zhaowei Li, Qi Xu, Dong Zhang, Hang Song, Yiqing Cai, Qi Qi, Ran Zhou, Junting Pan, Zefeng Li, Vu Tu, Zhida Huang, and Tao Wang.
\newblock Grounding{GPT}: Language enhanced multi-modal grounding model.
\newblock In \emph{ACL}, 2024{\natexlab{e}}.

\bibitem[Lin et~al.(2014)Lin, Maire, Belongie, Hays, Perona, Ramanan, Doll{\'a}r, and Zitnick]{lin2014microsoft}
Tsung-Yi Lin, Michael Maire, Serge Belongie, James Hays, Pietro Perona, Deva Ramanan, Piotr Doll{\'a}r, and C~Lawrence Zitnick.
\newblock Microsoft coco: Common objects in context.
\newblock In \emph{ECCV}, 2014.

\bibitem[Lin et~al.(2024)Lin, Liu, Zhang, Gao, Qiu, Xiao, Qiu, Shao, Chen, Han, Huang, Zhang, He, Qiao, and Li]{lin2023sphinx}
Ziyi Lin, Dongyang Liu, Renrui Zhang, Peng Gao, Longtian Qiu, Han Xiao, Han Qiu, Wenqi Shao, Keqin Chen, Jiaming Han, Siyuan Huang, Yichi Zhang, Xuming He, Yu Qiao, and Hongsheng Li.
\newblock Sphinx: A mixer of weights, visual embeddings and image scales for multi-modal large language models.
\newblock In \emph{ECCV}, 2024.

\bibitem[Liu et~al.(2023)Liu, Li, Wu, and Lee]{liu2024visual}
Haotian Liu, Chunyuan Li, Qingyang Wu, and Yong~Jae Lee.
\newblock Visual instruction tuning.
\newblock In \emph{NeurIPS}, 2023.

\bibitem[Liu et~al.(2024{\natexlab{a}})Liu, Zhang, Xu, Shi, Jiang, Yan, Zhang, Huang, Yuan, Li, and Hu]{liu2024mibench}
Haowei Liu, Xi Zhang, Haiyang Xu, Yaya Shi, Chaoya Jiang, Ming Yan, Ji Zhang, Fei Huang, Chunfeng Yuan, Bing Li, and Weiming Hu.
\newblock Mibench: Evaluating multimodal large language models over multiple images.
\newblock In \emph{EMNLP}, 2024{\natexlab{a}}.

\bibitem[Liu et~al.(2024{\natexlab{b}})Liu, Zeng, Ren, Li, Zhang, Yang, Li, Yang, Su, Zhu, and Zhang]{liu2023grounding}
Shilong Liu, Zhaoyang Zeng, Tianhe Ren, Feng Li, Hao Zhang, Jie Yang, Chunyuan Li, Jianwei Yang, Hang Su, Jun Zhu, and Lei Zhang.
\newblock Grounding dino: Marrying dino with grounded pre-training for open-set object detection.
\newblock In \emph{ECCV}, 2024{\natexlab{b}}.

\bibitem[Liu et~al.(2024{\natexlab{c}})Liu, Duan, Zhang, Li, Zhang, Zhao, Yuan, Wang, He, Liu, Chen, and Lin]{liu2023mmbench}
Yuan Liu, Haodong Duan, Yuanhan Zhang, Bo Li, Songyang Zhang, Wangbo Zhao, Yike Yuan, Jiaqi Wang, Conghui He, Ziwei Liu, Kai Chen, and Dahua Lin.
\newblock Mmbench: Is your multi-modal model an all-around player?
\newblock In \emph{ECCV}, 2024{\natexlab{c}}.

\bibitem[Liu et~al.(2021)Liu, Lin, Cao, Hu, Wei, Zhang, Lin, and Guo]{liu2021swin}
Ze Liu, Yutong Lin, Yue Cao, Han Hu, Yixuan Wei, Zheng Zhang, Stephen Lin, and Baining Guo.
\newblock Swin transformer: Hierarchical vision transformer using shifted windows.
\newblock In \emph{ICCV}, 2021.

\bibitem[Lu et~al.(2024)Lu, Quan, Zhu, and Yang]{lu2024zero}
Yu Lu, Ruijie Quan, Linchao Zhu, and Yi Yang.
\newblock Zero-shot video grounding with pseudo query lookup and verification.
\newblock \emph{IEEE TIP}, 2024.

\bibitem[Luo et~al.(2023)Luo, Zhao, Yang, Dong, Qiu, Lu, Wang, and Wei]{luo2023valley}
Ruipu Luo, Ziwang Zhao, Min Yang, Junwei Dong, Minghui Qiu, Pengcheng Lu, Tao Wang, and Zhongyu Wei.
\newblock Valley: Video assistant with large language model enhanced ability.
\newblock \emph{arXiv preprint arXiv:2306.07207}, 2023.

\bibitem[Ma et~al.(2024{\natexlab{a}})Ma, Jiang, Wu, Yuan, and Qi]{ma2024groma}
Chuofan Ma, Yi Jiang, Jiannan Wu, Zehuan Yuan, and Xiaojuan Qi.
\newblock Groma: Localized visual tokenization for grounding multimodal large language models.
\newblock In \emph{ECCV}, 2024{\natexlab{a}}.

\bibitem[Ma et~al.(2024{\natexlab{b}})Ma, Jin, Wang, Xian, Feng, and Yang]{ma2023vista}
Fan Ma, Xiaojie Jin, Heng Wang, Yuchen Xian, Jiashi Feng, and Yi Yang.
\newblock Vista-llama: Reliable video narrator via equal distance to visual tokens.
\newblock In \emph{CVPR}, 2024{\natexlab{b}}.

\bibitem[Mao et~al.(2016)Mao, Huang, Toshev, Camburu, Yuille, and Murphy]{mao2016generation}
Junhua Mao, Jonathan Huang, Alexander Toshev, Oana Camburu, Alan~L Yuille, and Kevin Murphy.
\newblock Generation and comprehension of unambiguous object descriptions.
\newblock In \emph{CVPR}, 2016.

\bibitem[Mathew et~al.(2021)Mathew, Karatzas, and Jawahar]{mathew2021docvqa}
Minesh Mathew, Dimosthenis Karatzas, and CV Jawahar.
\newblock Docvqa: A dataset for vqa on document images.
\newblock In \emph{WACV}, 2021.

\bibitem[Minderer et~al.(2022)Minderer, Gritsenko, Stone, Neumann, Weissenborn, Dosovitskiy, Mahendran, Arnab, Dehghani, Shen, et~al.]{minderer2022simple}
Matthias Minderer, Alexey Gritsenko, Austin Stone, Maxim Neumann, Dirk Weissenborn, Alexey Dosovitskiy, Aravindh Mahendran, Anurag Arnab, Mostafa Dehghani, Zhuoran Shen, et~al.
\newblock Simple open-vocabulary object detection.
\newblock In \emph{ECCV}, 2022.

\bibitem[Nguyen et~al.(2025)Nguyen, Juvekar, Yu, Wahed, and Lourentzou]{nguyen2025calico}
Kiet~A Nguyen, Adheesh Juvekar, Tianjiao Yu, Muntasir Wahed, and Ismini Lourentzou.
\newblock Calico: Part-focused semantic co-segmentation with large vision-language models.
\newblock In \emph{CVPR}, 2025.

\bibitem[Park et~al.(2019)Park, Darrell, and Rohrbach]{park2019robust}
Dong~Huk Park, Trevor Darrell, and Anna Rohrbach.
\newblock Robust change captioning.
\newblock In \emph{ICCV}, 2019.

\bibitem[Peng et~al.(2024)Peng, Wang, Dong, Hao, Huang, Ma, Ye, and Wei]{peng2023kosmos}
Zhiliang Peng, Wenhui Wang, Li Dong, Yaru Hao, Shaohan Huang, Shuming Ma, Qixiang Ye, and Furu Wei.
\newblock Grounding multimodal large language models to the world.
\newblock In \emph{ICLR}, 2024.

\bibitem[Pi et~al.(2023)Pi, Gao, Diao, Pan, Dong, Zhang, Yao, Han, Xu, Kong, and Zhang]{pi2023detgpt}
Renjie Pi, Jiahui Gao, Shizhe Diao, Rui Pan, Hanze Dong, Jipeng Zhang, Lewei Yao, Jianhua Han, Hang Xu, Lingpeng Kong, and Tong Zhang.
\newblock Detgpt: Detect what you need via reasoning.
\newblock In \emph{EMNLP}, 2023.

\bibitem[Qu et~al.(2023)Qu, Wu, Liu, Liang, Song, Zhao, and Wei]{qu2024rio}
Mengxue Qu, Yu Wu, Wu Liu, Xiaodan Liang, Jingkuan Song, Yao Zhao, and Yunchao Wei.
\newblock Rio: A benchmark for reasoning intention-oriented objects in open environments.
\newblock In \emph{NeurIPS}, 2023.

\bibitem[Radford et~al.(2021)Radford, Kim, Hallacy, Ramesh, Goh, Agarwal, Sastry, Askell, Mishkin, Clark, et~al.]{radford2021learning}
Alec Radford, Jong~Wook Kim, Chris Hallacy, Aditya Ramesh, Gabriel Goh, Sandhini Agarwal, Girish Sastry, Amanda Askell, Pamela Mishkin, Jack Clark, et~al.
\newblock Learning transferable visual models from natural language supervision.
\newblock In \emph{ICML}, 2021.

\bibitem[Rasheed et~al.(2024)Rasheed, Maaz, Shaji, Shaker, Khan, Cholakkal, Anwer, Xing, Yang, and Khan]{rasheed2023glamm}
Hanoona Rasheed, Muhammad Maaz, Sahal Shaji, Abdelrahman Shaker, Salman Khan, Hisham Cholakkal, Rao~M Anwer, Erix Xing, Ming-Hsuan Yang, and Fahad~S Khan.
\newblock Glamm: Pixel grounding large multimodal model.
\newblock In \emph{CVPR}, 2024.

\bibitem[Ravi et~al.(2025)Ravi, Gabeur, Hu, Hu, Ryali, Ma, Khedr, R{\"a}dle, Rolland, Gustafson, et~al.]{ravi2024sam}
Nikhila Ravi, Valentin Gabeur, Yuan-Ting Hu, Ronghang Hu, Chaitanya Ryali, Tengyu Ma, Haitham Khedr, Roman R{\"a}dle, Chloe Rolland, Laura Gustafson, et~al.
\newblock Sam 2: Segment anything in images and videos.
\newblock In \emph{ICLR}, 2025.

\bibitem[Redmon et~al.(2016)Redmon, Divvala, Girshick, and Farhadi]{redmon2016you}
Joseph Redmon, Santosh Divvala, Ross Girshick, and Ali Farhadi.
\newblock You only look once: Unified, real-time object detection.
\newblock In \emph{CVPR}, 2016.

\bibitem[Reid et~al.(2024)Reid, Savinov, Teplyashin, Lepikhin, Lillicrap, Alayrac, Soricut, Lazaridou, Firat, Schrittwieser, et~al.]{reid2024gemini}
Machel Reid, Nikolay Savinov, Denis Teplyashin, Dmitry Lepikhin, Timothy Lillicrap, Jean-baptiste Alayrac, Radu Soricut, Angeliki Lazaridou, Orhan Firat, Julian Schrittwieser, et~al.
\newblock Gemini 1.5: Unlocking multimodal understanding across millions of tokens of context.
\newblock \emph{arXiv preprint arXiv:2403.05530}, 2024.

\bibitem[Ren et~al.(2015)Ren, He, Girshick, and Sun]{ren2015faster}
Shaoqing Ren, Kaiming He, Ross Girshick, and Jian Sun.
\newblock Faster r-cnn: Towards real-time object detection with region proposal networks.
\newblock In \emph{NeurIPS}, 2015.

\bibitem[Ren et~al.(2024)Ren, Yao, Li, Sun, and Hou]{ren2024timechat}
Shuhuai Ren, Linli Yao, Shicheng Li, Xu Sun, and Lu Hou.
\newblock Timechat: A time-sensitive multimodal large language model for long video understanding.
\newblock In \emph{CVPR}, 2024.

\bibitem[Schulter et~al.(2023)Schulter, Suh, Dafnis, Zhang, Zhao, Metaxas, et~al.]{schulter2023omnilabel}
Samuel Schulter, Yumin Suh, Konstantinos~M Dafnis, Zhixing Zhang, Shiyu Zhao, Dimitris Metaxas, et~al.
\newblock {OmniLabel}: A challenging benchmark for language-based object detection.
\newblock In \emph{ICCV}, 2023.

\bibitem[Shao et~al.(2024)Shao, Hu, Wang, Song, Waslander, Liu, and Li]{shao2024lmdrive}
Hao Shao, Yuxuan Hu, Letian Wang, Guanglu Song, Steven~L Waslander, Yu Liu, and Hongsheng Li.
\newblock Lmdrive: Closed-loop end-to-end driving with large language models.
\newblock In \emph{CVPR}, 2024.

\bibitem[Shen et~al.(2024)Shen, Fu, Chen, Zhang, Li, Sun, Wu, Lin, and Ji]{APE}
Yunhang Shen, Chaoyou Fu, Peixian Chen, Mengdan Zhang, Ke Li, Xing Sun, Yunsheng Wu, Shaohui Lin, and Rongrong Ji.
\newblock Aligning and prompting everything all at once for universal visual perception.
\newblock In \emph{CVPR}, 2024.

\bibitem[Song et~al.(2024)Song, Chen, Chen, Yu, Wan, and Wang]{song2024milebench}
Dingjie Song, Shunian Chen, Guiming~Hardy Chen, Fei Yu, Xiang Wan, and Benyou Wang.
\newblock Milebench: Benchmarking mllms in long context.
\newblock In \emph{COLM}, 2024.

\bibitem[Subramanian et~al.(2022)Subramanian, Merrill, Darrell, Gardner, Singh, and Rohrbach]{subramanian2022reclip}
Sanjay Subramanian, William Merrill, Trevor Darrell, Matt Gardner, Sameer Singh, and Anna Rohrbach.
\newblock Reclip: A strong zero-shot baseline for referring expression comprehension.
\newblock In \emph{ACL}, 2022.

\bibitem[Suhr et~al.(2019)Suhr, Zhou, Zhang, Zhang, Bai, and Artzi]{suhr2019corpus}
Alane Suhr, Stephanie Zhou, Ally Zhang, Iris Zhang, Huajun Bai, and Yoav Artzi.
\newblock A corpus for reasoning about natural language grounded in photographs.
\newblock In \emph{ACL}, 2019.

\bibitem[Tanaka et~al.(2023)Tanaka, Nishida, Nishida, Hasegawa, Saito, and Saito]{tanaka2023slidevqa}
Ryota Tanaka, Kyosuke Nishida, Kosuke Nishida, Taku Hasegawa, Itsumi Saito, and Kuniko Saito.
\newblock Slidevqa: A dataset for document visual question answering on multiple images.
\newblock In \emph{AAAI}, 2023.

\bibitem[Tang et~al.(2014)Tang, Joulin, Li, and Fei-Fei]{tang2014co}
Kevin Tang, Armand Joulin, Li-Jia Li, and Li Fei-Fei.
\newblock Co-localization in real-world images.
\newblock In \emph{CVPR}, 2014.

\bibitem[Tkachenko et~al.(2020-2022)Tkachenko, Malyuk, Holmanyuk, and Liubimov]{LabelStudio}
Maxim Tkachenko, Mikhail Malyuk, Andrey Holmanyuk, and Nikolai Liubimov.
\newblock {Label Studio}: Data labeling software, 2020-2022.
\newblock Open source software available from \url{https://github.com/heartexlabs/label-studio}.

\bibitem[Wahed et~al.(2024)Wahed, Nguyen, Juvekar, Li, Zhou, Shah, Yu, Yanardag, and Lourentzou]{wahed2024prima}
Muntasir Wahed, Kiet~A Nguyen, Adheesh~Sunil Juvekar, Xinzhuo Li, Xiaona Zhou, Vedant Shah, Tianjiao Yu, Pinar Yanardag, and Ismini Lourentzou.
\newblock Prima: Multi-image vision-language models for reasoning segmentation.
\newblock \emph{arXiv preprint arXiv:2412.15209}, 2024.

\bibitem[Wang et~al.(2025)Wang, Fan, Quan, Yao, and Yang]{wang2024protchatgpt}
Chao Wang, Hehe Fan, Ruijie Quan, Lina Yao, and Yi Yang.
\newblock Protchatgpt: Towards understanding proteins with large language models.
\newblock In \emph{SIGIR}, 2025.

\bibitem[Wang et~al.(2024{\natexlab{a}})Wang, Zhan, Liu, Ding, and Yu]{wang2024balanced}
Hanyao Wang, Yibing Zhan, Liu Liu, Liang Ding, and Jun Yu.
\newblock Balanced similarity with auxiliary prompts: Towards alleviating text-to-image retrieval bias for clip in zero-shot learning.
\newblock \emph{arXiv preprint arXiv:2402.18400}, 2024{\natexlab{a}}.

\bibitem[Wang et~al.(2023)Wang, Wang, Lin, Bai, Zhou, Zhou, Wang, and Zhou]{wang2023one}
Peng Wang, Shijie Wang, Junyang Lin, Shuai Bai, Xiaohuan Zhou, Jingren Zhou, Xinggang Wang, and Chang Zhou.
\newblock One-peace: Exploring one general representation model toward unlimited modalities.
\newblock \emph{arXiv preprint arXiv:2305.11172}, 2023.

\bibitem[Wang et~al.(2024{\natexlab{b}})Wang, Bai, Tan, Wang, Fan, Bai, Chen, Liu, Wang, Ge, Fan, Dang, Du, Ren, Men, Liu, Zhou, Zhou, and Lin]{Qwen2-VL}
Peng Wang, Shuai Bai, Sinan Tan, Shijie Wang, Zhihao Fan, Jinze Bai, Keqin Chen, Xuejing Liu, Jialin Wang, Wenbin Ge, Yang Fan, Kai Dang, Mengfei Du, Xuancheng Ren, Rui Men, Dayiheng Liu, Chang Zhou, Jingren Zhou, and Junyang Lin.
\newblock Qwen2-vl: Enhancing vision-language model's perception of the world at any resolution.
\newblock \emph{arXiv preprint arXiv:2409.12191}, 2024{\natexlab{b}}.

\bibitem[Wang et~al.(2024{\natexlab{c}})Wang, Lv, Yu, Hong, Qi, Wang, Ji, Yang, Zhao, XiXuan, Xu, Chen, Xu, Li, Dong, Ding, and Tang]{wang2023cogvlm}
Weihan Wang, Qingsong Lv, Wenmeng Yu, Wenyi Hong, Ji Qi, Yan Wang, Junhui Ji, Zhuoyi Yang, Lei Zhao, Song XiXuan, Jiazheng Xu, Keqin Chen, Bin Xu, Juanzi Li, Yuxiao Dong, Ming Ding, and Jie Tang.
\newblock Cogvlm: Visual expert for pretrained language models.
\newblock In \emph{NeurIPS}, 2024{\natexlab{c}}.

\bibitem[Wang et~al.(2024{\natexlab{d}})Wang, Zhou, Liu, Lu, Xu, He, Yoon, Lu, Liu, Bertasius, Bansal, Yao, and Huang]{wang2024mementos}
Xiyao Wang, Yuhang Zhou, Xiaoyu Liu, Hongjin Lu, Yuancheng Xu, Feihong He, Jaehong Yoon, Taixi Lu, Fuxiao Liu, Gedas Bertasius, Mohit Bansal, Huaxiu Yao, and Furong Huang.
\newblock Mementos: A comprehensive benchmark for multimodal large language model reasoning over image sequences.
\newblock In \emph{ACL}, 2024{\natexlab{d}}.

\bibitem[Wei et~al.(2025)Wei, Zhang, Zhang, Zhang, and Chu]{wei2023lenna}
Fei Wei, Xinyu Zhang, Ailing Zhang, Bo Zhang, and Xiangxiang Chu.
\newblock Lenna: Language enhanced reasoning detection assistant.
\newblock In \emph{ICASSP}, 2025.

\bibitem[Wei et~al.(2022)Wei, Wang, Schuurmans, Bosma, Xia, Chi, Le, Zhou, et~al.]{wei2022chain}
Jason Wei, Xuezhi Wang, Dale Schuurmans, Maarten Bosma, Fei Xia, Ed~H Chi, Quoc~V Le, Denny Zhou, et~al.
\newblock Chain-of-thought prompting elicits reasoning in large language models.
\newblock In \emph{NeurIPS}, 2022.

\bibitem[Wu et~al.(2021)Wu, Yu, Chen, Tenenbaum, and Gan]{wu2024star}
Bo Wu, Shoubin Yu, Zhenfang Chen, Joshua~B Tenenbaum, and Chuang Gan.
\newblock Star: A benchmark for situated reasoning in real-world videos.
\newblock In \emph{NeurIPS}, 2021.

\bibitem[Wu et~al.(2024{\natexlab{a}})Wu, Zhang, Zhang, Chen, Liao, Wang, Li, Sun, Yan, Zhai, and Lin]{wu2024qbench}
Haoning Wu, Zicheng Zhang, Erli Zhang, Chaofeng Chen, Liang Liao, Annan Wang, Chunyi Li, Wenxiu Sun, Qiong Yan, Guangtao Zhai, and Weisi Lin.
\newblock Q-bench: A benchmark for general-purpose foundation models on low-level vision.
\newblock In \emph{ICLR}, 2024{\natexlab{a}}.

\bibitem[Wu et~al.(2024{\natexlab{b}})Wu, Zhu, Zhang, Zhang, Chen, Liao, Li, Wang, Sun, Yan, et~al.]{wu2024towards}
Haoning Wu, Hanwei Zhu, Zicheng Zhang, Erli Zhang, Chaofeng Chen, Liang Liao, Chunyi Li, Annan Wang, Wenxiu Sun, Qiong Yan, et~al.
\newblock Towards open-ended visual quality comparison.
\newblock In \emph{ECCV}, 2024{\natexlab{b}}.

\bibitem[Wu et~al.(2023)Wu, Zhang, Xie, Zhu, and Zhao]{wu2023advancing}
Yixuan Wu, Zhao Zhang, Chi Xie, Feng Zhu, and Rui Zhao.
\newblock Advancing referring expression segmentation beyond single image.
\newblock In \emph{ICCV}, 2023.

\bibitem[Wu et~al.(2025)Wu, Chen, Li, Fan, and Yang]{wu2025bvinet}
Zhiliang Wu, Kerui Chen, Kun Li, Hehe Fan, and Yi Yang.
\newblock Bvinet: Unlocking blind video inpainting with zero annotations.
\newblock In \emph{ICCV}, 2025.

\bibitem[Xiao et~al.(2024)Xiao, Wu, Xu, Dai, Hu, Lu, Zeng, Liu, and Yuan]{xiao2024florence}
Bin Xiao, Haiping Wu, Weijian Xu, Xiyang Dai, Houdong Hu, Yumao Lu, Michael Zeng, Ce Liu, and Lu Yuan.
\newblock Florence-2: Advancing a unified representation for a variety of vision tasks.
\newblock In \emph{CVPR}, 2024.

\bibitem[Xie et~al.(2023)Xie, Zhang, Wu, Zhu, Zhao, and Liang]{NEURIPS2023_f9fd24fd}
Chi Xie, Zhao Zhang, Yixuan Wu, Feng Zhu, Rui Zhao, and Shuang Liang.
\newblock Described object detection: Liberating object detection with flexible expressions.
\newblock In \emph{NeurIPS}, 2023.

\bibitem[Xu et~al.(2021)Xu, Zhou, Yu, Xiao, and Yang]{xu2021pyramidal}
Yunqiu Xu, Chunluan Zhou, Xin Yu, Bin Xiao, and Yi Yang.
\newblock Pyramidal multiple instance detection network with mask guided self-correction for weakly supervised object detection.
\newblock \emph{IEEE TIP}, 2021.

\bibitem[Xu et~al.(2022)Xu, Sun, Yang, Miao, and Yang]{xu2022h2fa}
Yunqiu Xu, Yifan Sun, Zongxin Yang, Jiaxu Miao, and Yi Yang.
\newblock H2fa r-cnn: Holistic and hierarchical feature alignment for cross-domain weakly supervised object detection.
\newblock In \emph{CVPR}, 2022.

\bibitem[Xu et~al.(2023)Xu, Zhou, Yu, and Yang]{xu2023cyclic}
Yunqiu Xu, Chunluan Zhou, Xin Yu, and Yi Yang.
\newblock Cyclic self-training with proposal weight modulation for cross-supervised object detection.
\newblock \emph{IEEE TIP}, 2023.

\bibitem[Xu et~al.(2024)Xu, Zhu, and Yang]{xu2024gg}
Yunqiu Xu, Linchao Zhu, and Yi Yang.
\newblock Gg-editor: Locally editing 3d avatars with multimodal large language model guidance.
\newblock In \emph{ACM MM}, 2024.

\bibitem[Yan et~al.(2023)Yan, Yang, Zhu, Lin, Li, Wang, Yang, Zhong, McAuley, Gao, et~al.]{yan2023gpt}
An Yan, Zhengyuan Yang, Wanrong Zhu, Kevin Lin, Linjie Li, Jianfeng Wang, Jianwei Yang, Yiwu Zhong, Julian McAuley, Jianfeng Gao, et~al.
\newblock {GPT-4V} in wonderland: Large multimodal models for zero-shot smartphone gui navigation.
\newblock \emph{arXiv preprint arXiv:2311.07562}, 2023.

\bibitem[You et~al.(2024)You, Zhang, Gan, Du, Zhang, Wang, Cao, Chang, and Yang]{you2023ferret}
Haoxuan You, Haotian Zhang, Zhe Gan, Xianzhi Du, Bowen Zhang, Zirui Wang, Liangliang Cao, Shih-Fu Chang, and Yinfei Yang.
\newblock Ferret: Refer and ground anything anywhere at any granularity.
\newblock In \emph{ICLR}, 2024.

\bibitem[Yue et~al.(2024)Yue, Ni, Zhang, Zheng, Liu, Zhang, Stevens, Jiang, Ren, Sun, et~al.]{yue2023mmmu}
Xiang Yue, Yuansheng Ni, Kai Zhang, Tianyu Zheng, Ruoqi Liu, Ge Zhang, Samuel Stevens, Dongfu Jiang, Weiming Ren, Yuxuan Sun, et~al.
\newblock Mmmu: A massive multi-discipline multimodal understanding and reasoning benchmark for expert agi.
\newblock In \emph{CVPR}, 2024.

\bibitem[Zareian et~al.(2021)Zareian, Rosa, Hu, and Chang]{zareian2021open}
Alireza Zareian, Kevin~Dela Rosa, Derek~Hao Hu, and Shih-Fu Chang.
\newblock Open-vocabulary object detection using captions.
\newblock In \emph{CVPR}, 2021.

\bibitem[Zhan et~al.(2024)Zhan, Zhu, Chen, Yang, Tang, and Wang]{zhan2025griffonv1}
Yufei Zhan, Yousong Zhu, Zhiyang Chen, Fan Yang, Ming Tang, and Jinqiao Wang.
\newblock Griffon: Spelling out all object locations at any granularity with large language models.
\newblock In \emph{ECCV}, 2024.

\bibitem[Zhang et~al.(2022)Zhang, Zhang, Hu, Chen, Li, Dai, Wang, Yuan, Hwang, and Gao]{NEURIPS2022_ea370419}
Haotian Zhang, Pengchuan Zhang, Xiaowei Hu, Yen-Chun Chen, Liunian Li, Xiyang Dai, Lijuan Wang, Lu Yuan, Jenq-Neng Hwang, and Jianfeng Gao.
\newblock Glipv2: Unifying localization and vision-language understanding.
\newblock In \emph{NeurIPS}, 2022.

\bibitem[Zhang et~al.(2023)Zhang, Li, and Bing]{zhang2023video}
Hang Zhang, Xin Li, and Lidong Bing.
\newblock Video-llama: An instruction-tuned audio-visual language model for video understanding.
\newblock In \emph{EMNLP}, 2023.

\bibitem[Zhang et~al.(2024{\natexlab{a}})Zhang, Li, Li, Ren, Zou, Liu, Huang, Gao, Zhang, Li, and Yang]{zhang2023llava}
Hao Zhang, Hongyang Li, Feng Li, Tianhe Ren, Xueyan Zou, Shilong Liu, Shijia Huang, Jianfeng Gao, Lei Zhang, Chunyuan Li, and Jianwei Yang.
\newblock Llava-grounding: Grounded visual chat with large multimodal models.
\newblock In \emph{ECCV}, 2024{\natexlab{a}}.

\bibitem[Zhang et~al.(2024{\natexlab{b}})Zhang, You, Dufter, Zhang, Chen, Chen, Fu, Wang, Chang, Gan, et~al.]{zhang2024ferret}
Haotian Zhang, Haoxuan You, Philipp Dufter, Bowen Zhang, Chen Chen, Hong-You Chen, Tsu-Jui Fu, William~Yang Wang, Shih-Fu Chang, Zhe Gan, et~al.
\newblock Ferret-v2: An improved baseline for referring and grounding with large language models.
\newblock In \emph{COLM}, 2024{\natexlab{b}}.

\bibitem[Zhang et~al.(2024{\natexlab{c}})Zhang, Ma, Gao, Shakiah, Gao, and Chai]{zhang2024groundhog}
Yichi Zhang, Ziqiao Ma, Xiaofeng Gao, Suhaila Shakiah, Qiaozi Gao, and Joyce Chai.
\newblock Groundhog: Grounding large language models to holistic segmentation.
\newblock In \emph{CVPR}, 2024{\natexlab{c}}.

\bibitem[Zhang et~al.(2025)Zhang, Lu, Wang, Rao, Yang, and Zhu]{zhang2025flexselect}
Yunzhu Zhang, Yu Lu, Tianyi Wang, Fengyun Rao, Yi Yang, and Linchao Zhu.
\newblock Flexselect: Flexible token selection for efficient long video understanding.
\newblock \emph{arXiv preprint arXiv:2506.00993}, 2025.

\bibitem[Zhao et~al.(2023)Zhao, Lin, Zhou, Huang, Feng, and Kang]{zhao2023bubogpt}
Yang Zhao, Zhijie Lin, Daquan Zhou, Zilong Huang, Jiashi Feng, and Bingyi Kang.
\newblock Bubogpt: Enabling visual grounding in multi-modal llms.
\newblock \emph{arXiv preprint arXiv:2307.08581}, 2023.

\bibitem[Zheng et~al.(2024)Zheng, Zhang, Zhang, Ye, Luo, Feng, and Ma]{zheng2024llamafactory}
Yaowei Zheng, Richong Zhang, Junhao Zhang, Yanhan Ye, Zheyan Luo, Zhangchi Feng, and Yongqiang Ma.
\newblock Llamafactory: Unified efficient fine-tuning of 100+ language models.
\newblock In \emph{ACL}, 2024.

\bibitem[Zhu et~al.(2023)Zhu, Chen, Shen, Li, and Elhoseiny]{zhu2023minigpt}
Deyao Zhu, Jun Chen, Xiaoqian Shen, Xiang Li, and Mohamed Elhoseiny.
\newblock Minigpt-4: Enhancing vision-language understanding with advanced large language models.
\newblock In \emph{ICLR}, 2023.

\bibitem[Zhu et~al.(2025)Zhu, Jia, Zhang, Li, and Jiang]{zhu2024multichartqa}
Zifeng Zhu, Mengzhao Jia, Zhihan Zhang, Lang Li, and Meng Jiang.
\newblock Multichartqa: Benchmarking vision-language models on multi-chart problems.
\newblock In \emph{NAACL}, 2025.

\end{thebibliography}
}

\end{document}